\documentclass[sigconf]{acmart}

\AtBeginDocument{%
  }

\copyrightyear{2025} 
\acmYear{2025} 
\setcopyright{acmlicensed}\acmConference[KDD '25]{Proceedings of the 31st ACM
SIGKDD Conference on Knowledge Discovery and Data Mining V.1}{August 3--7,
2025}{Toronto, ON, Canada}
\acmBooktitle{Proceedings of the 31st ACM SIGKDD Conference on Knowledge
Discovery and Data Mining V.1 (KDD '25), August 3--7, 2025, Toronto, ON, Canada}
\acmDOI{10.1145/3690624.3709325}
\acmISBN{979-8-4007-1245-6/25/08}

\usepackage{makecell}
\usepackage{array}
\usepackage{algorithmicx,algorithm}
\usepackage{multirow}
\usepackage[normalem]{ulem}
\usepackage{subcaption}
\usepackage{graphicx}
\usepackage{tabularx}
\usepackage{color}
\usepackage{xspace} 
\usepackage[utf8]{inputenc}
\usepackage[T1]{fontenc}
\usepackage{CJKutf8}
\usepackage{enumitem}
\usepackage{diagbox}
\usepackage{balance}
\usepackage[flushleft]{threeparttable}
\usepackage{marvosym}

\newcommand{\icon}{\raisebox{-1pt}{\includegraphics[width=1.0em]{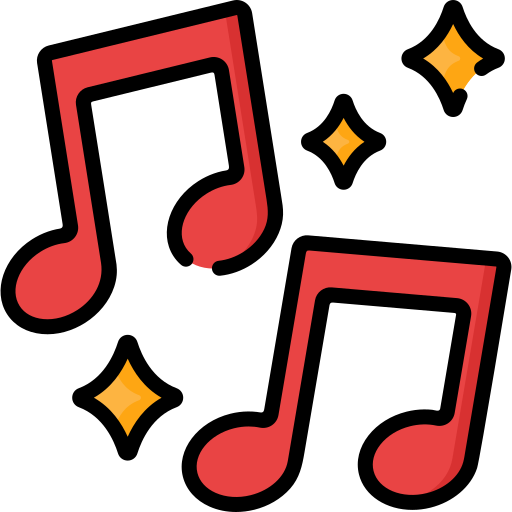}}\xspace}

\begin{document}

\title{\icon ~~DUET: Dual Clustering Enhanced Multivariate Time \\Series Forecasting}


\author{Xiangfei Qiu}
\affiliation{%
  \institution{East China Normal University\textsuperscript{1}}
  \city{Shanghai}
  \country{China}
}
\email{xfqiu@stu.ecnu.edu.cn}

\author{Xingjian Wu}
\affiliation{%
  \institution{East China Normal University\textsuperscript{1}}%
  \city{Shanghai}
  \country{China}
}
\email{xjwu@stu.ecnu.edu.cn}

\author{Yan Lin}
\affiliation{%
  \institution{Aalborg University}
  \city{Aalborg}
  \country{Denmark}
}
\email{lyan@cs.aau.dk}

\author{Chenjuan Guo}
\affiliation{%
  \institution{East China Normal University\textsuperscript{1}}
    \city{Shanghai}
  \country{China}
}
\email{cjguo@dase.ecnu.edu.cn}

\author{Jilin Hu \Letter}
\affiliation{%
  \institution{East China Normal University\textsuperscript{1,}\textsuperscript{2}}
    \city{Shanghai}
  \country{China}
}
\email{jlhu@dase.ecnu.edu.cn}

\author{Bin Yang}
\affiliation{%
  \institution{East China Normal University\textsuperscript{1}}
    \city{Shanghai}
  \country{China}
}
\email{byang@dase.ecnu.edu.cn}

\renewcommand{\shortauthors}{Xiangfei Qiu et al.}

\begin{abstract}
Multivariate time series forecasting is crucial for various applications, such as financial investment, energy management, weather forecasting, and traffic optimization. However, accurate forecasting is challenging due to two main factors. First, real-world time series often show heterogeneous temporal patterns caused by distribution shifts over time. Second, correlations among channels are complex and intertwined, making it hard to model the interactions among channels precisely and flexibly.

In this study, we address these challenges by proposing a general framework called \textbf{DUET}, which introduces \underline{DU}al clustering on the temporal and channel dimensions to \underline{E}nhance multivariate \underline{T}ime series forecasting. First, we design a Temporal Clustering Module (TCM) that clusters time series into fine-grained distributions to handle heterogeneous temporal patterns. For different distribution clusters, we design various pattern extractors to capture their intrinsic temporal patterns, thus modeling the heterogeneity. Second, we introduce a novel Channel-Soft-Clustering strategy and design a Channel Clustering Module (CCM), which captures the relationships among channels in the frequency domain through metric learning and applies sparsification to mitigate the adverse effects of noisy channels. Finally, DUET combines TCM and CCM to incorporate both the temporal and channel dimensions. Extensive experiments on 25 real-world datasets from 10 application domains, demonstrate the state-of-the-art performance of DUET. 
\footnotetext[1]{School of Data Science \& Engineering.}
\footnotetext[2]{Engineering Research Center of Blockchain Data Management, Ministry of Education.}
\end{abstract}





\maketitle

\section{Introduction}

\begin{figure}[h!]
  \centering
  \includegraphics[width=0.9\linewidth]{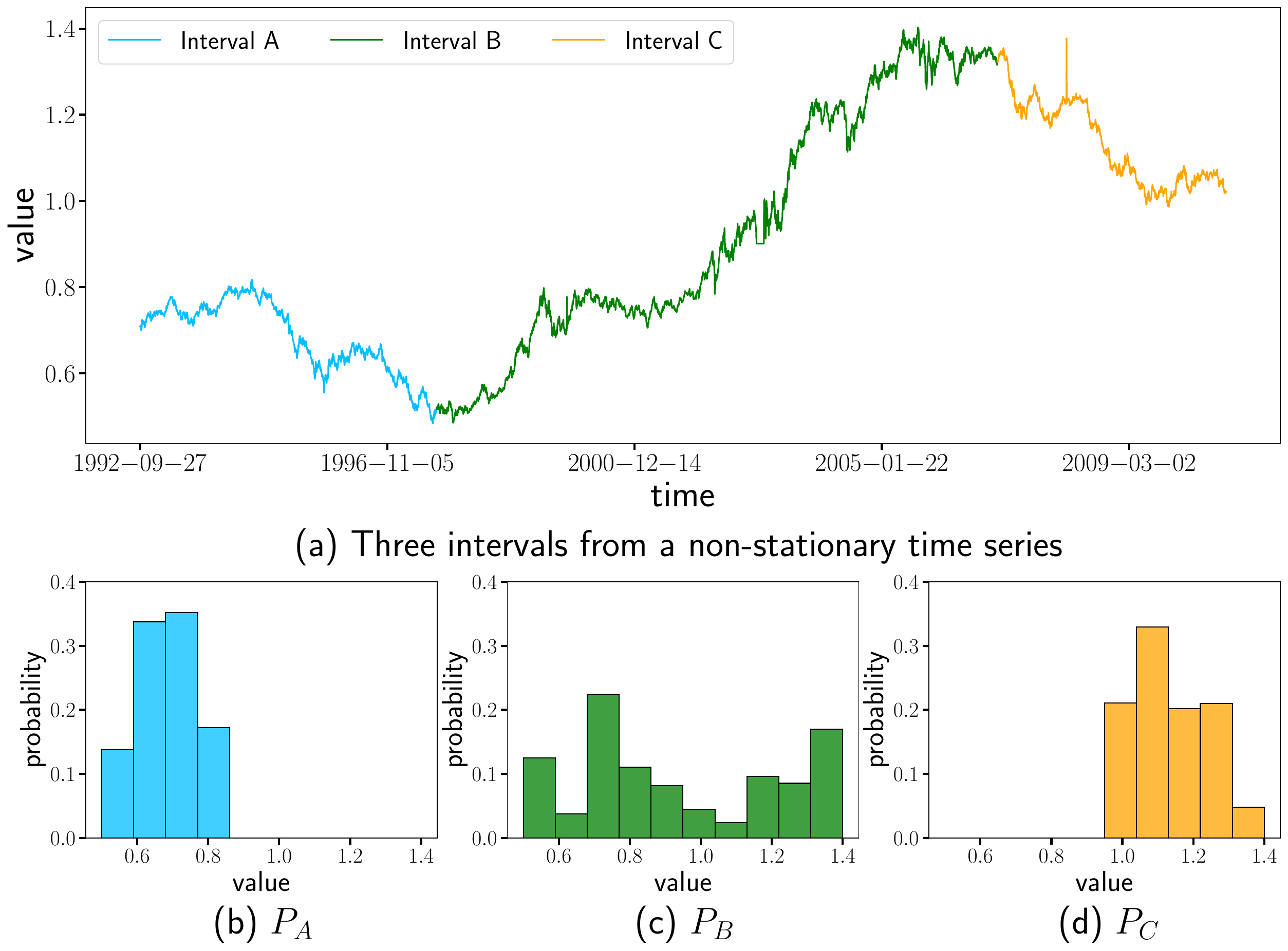}
    \vspace{-3mm}
  \caption{A Non-stationary time series with three intervals A, B, and 
 C, exhibiting varying value distributions ($P_A \ne P_B \ne P_C$) and temporal patterns.}
  \label{shifting show}
  \vspace{-3mm}
\end{figure}

Multivariate time series is a type of time series that organizes timestamps chronologically and involves multiple channels~(a.k.a., variables) at each timestamp~\cite{DBLP:conf/ijcai/HuangZYLYW24,HaoWang4,wangrethinking,hu2024attractor,hu2024time,sun2024hierarchical}.
In recent years, multivariate time series analysis has seen remarkable progress, with key tasks such as anomaly detection~\cite{D3R, liu2024time, liu2024elephant, DCdetector, hu2024multirc, wu2024catch}, classification~\cite{DBLP:conf/icde/YaoJC0GW24,DBLP:journals/pacmmod/0002Z0KGJ23}, and imputation~\cite{gao2024diffimp,wang2024spot,escmtifs,lsptd}, among others~\cite{DBLP:conf/nips/HuangSZDWZW23,DBLP:journals/pvldb/YaoDLCLGL24,DBLP:journals/pvldb/YaoLJ00CW0G23,hu2024toward}, gaining attention. Among these, multivariate time series forecasting~(MTSF)~\cite{yu2024ginar,DSformer,wang2024rose,zhu2024fcnet,HybridZheng,DBLP:conf/aaai/HuangSZCDZW24,DBLP:journals/pvldb/ChengCGZWYJ23} stands out as a critical and widely studied task. It has been extensively applied in diverse domains, including economics~\cite{sezer2020financial,huang2022dgraph}, traffic~\cite{wu2024autocts++,wu2024fully,cirstea2022towards,DBLP:journals/pvldb/FangPCDG21,kieu2024Team,DBLP:conf/ijcai/YangGHT021,DBLP:journals/tkde/YangGY22}, energy~\cite{HaoWang1,guo2015ecomark,sun2022solar}, and AIOps~\cite{lin2024cocv,davidpvldb,Monotonic,pan2023magicscaler,lin2025benchmarking}, highlighting its importance and impact.

Building an MTSF method typically involves modeling correlations on the temporal and channel dimensions. However, real-world time series often exhibit heterogeneous temporal patterns caused by the shifting of distribution over time, a phenomenon called \textit{Temporal Distribution Shift} (\textbf{TDS})~\cite{brockwell1991time, du2021adarnn}. Additionally, the correlation among multiple channels is complex and intertwined~\cite{qiu2024tfb}. Therefore, developing a method that can effectively extract heterogeneous temporal patterns and channel dependencies is essential yet challenging. Specifically, achieving these goals is hindered by two major challenges.

\textbf{First, heterogeneous temporal patterns caused by TDS are difficult to model.}
In real applications, time series that describe unstable systems are easily influenced by external factors~\cite{nason2006stationary, young1999nonstationary}. Such non-stationarity of time series implies the data distribution changes over time, a phenomenon called TDS~\cite{du2021adarnn, fan2023dish, brockwell1991time}. TDS causes time series to have different temporal patterns, formally known as heterogeneity of temporal patterns~\cite{du2021adarnn, fan2023dish, lu2024diversify, lu2022out}. For example, Figure~\ref{shifting show}(a) illustrates a time series in economics, which shows the changes with the international circumstances. We can observe that the three intervals A, B, and C follow different temporal distributions, as evidenced by the value histograms shown in Figures~\ref{shifting show}(b), \ref{shifting show}(c), and~\ref{shifting show}(d). This shift in distribution also comes with varying temporal patterns. As shown in Figure~\ref{shifting show}(a), the blue interval A shows a descending trend, the green interval B shows an increasing trend, and the yellow interval C shows a steeper descending trend. 
It is crucial to incorporate these patterns considering their common presence in time series. However, recent studies~\cite{wang2024timemixer,liu2023itransformer,xu2024fitsmodelingtimeseries,lin2024sparsetsf} primarily address heterogeneous temporal patterns in an implicit manner, which ultimately undermines prediction accuracy.

\begin{figure}[t!]
  \centering
    \includegraphics[width=1\linewidth]{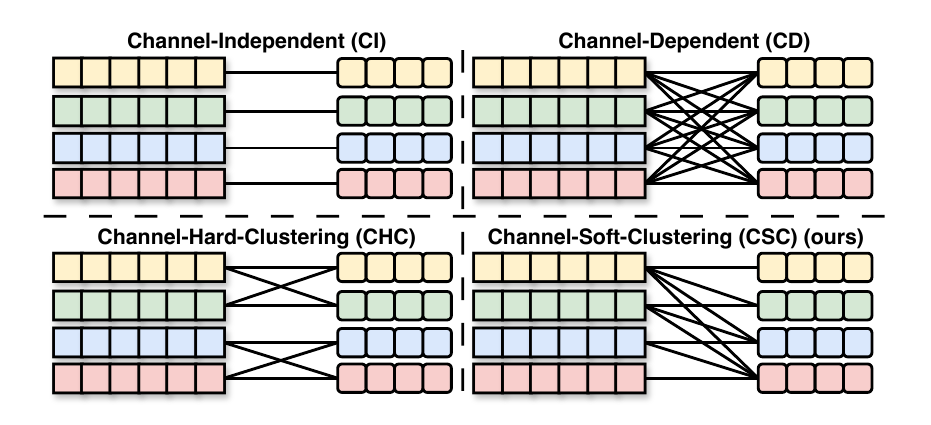}
      \vspace{-5mm}
  \caption{Channel strategies. Different colors represent different channels, with squares representing features before processing with various channel strategies, and squares with rounded corners representing features after processing.}
  \vspace{-3mm}
  \label{channel strategy}
\end{figure}

\textbf{Second, complex channel interrelations are difficult to model flexibly. }
For an MTSF task, it is crucial to model the correlations among different channels, as the predictive accuracy for a particular channel can often be enhanced by leveraging information from other related channels. For example, in weather forecasting, temperature predictions can be improved by incorporating data on humidity, wind speed, and pressure, as these factors are interrelated and provide a more comprehensive view of weather conditions.

Researchers have explored various strategies to manage multiple channels, including 1) treating each channel independently~\cite{nie2022time, liu2023itransformer}~(\textit{Channel-Independent}, \textbf{CI}), 2) assuming each channel correlates with all other channels~\cite{zhang2022crossformer}~(\textit{Channel-Dependent}, \textbf{CD}), and 3) grouping channels into clusters~\cite{liu2024dgcformer}~(\textit{Channel-Hard-Clustering}, \textbf{CHC}). Figure~\ref{channel strategy} illustrates these three strategies. CI imposes the constraint of using the same model across different channels. While it offers robustness~\cite{nie2022time}, it overlooks potential interactions among channels and can be limited in generalizability and capacity for unseen channels~\cite{qiu2024tfb, han2024capacity}. CD, on the other hand, considers all channels simultaneously and generates joint representations for decoding~\cite{zhang2022crossformer}, but may be susceptible to noise from irrelevant channels, reducing the model's robustness. CHC partitions multivariate time series into disjoint clusters through hard clustering, applying CD modeling methods within each cluster and CI methods among clusters~\cite{liu2024dgcformer}. However, this approach only considers relationships within the same cluster, limiting flexibility and versatility. 
In conclusion, there is yet an approach to model the complex interactions among channels precisely and flexibly.

\begin{figure}[t!]
  \centering
  \includegraphics[width=0.7\linewidth]{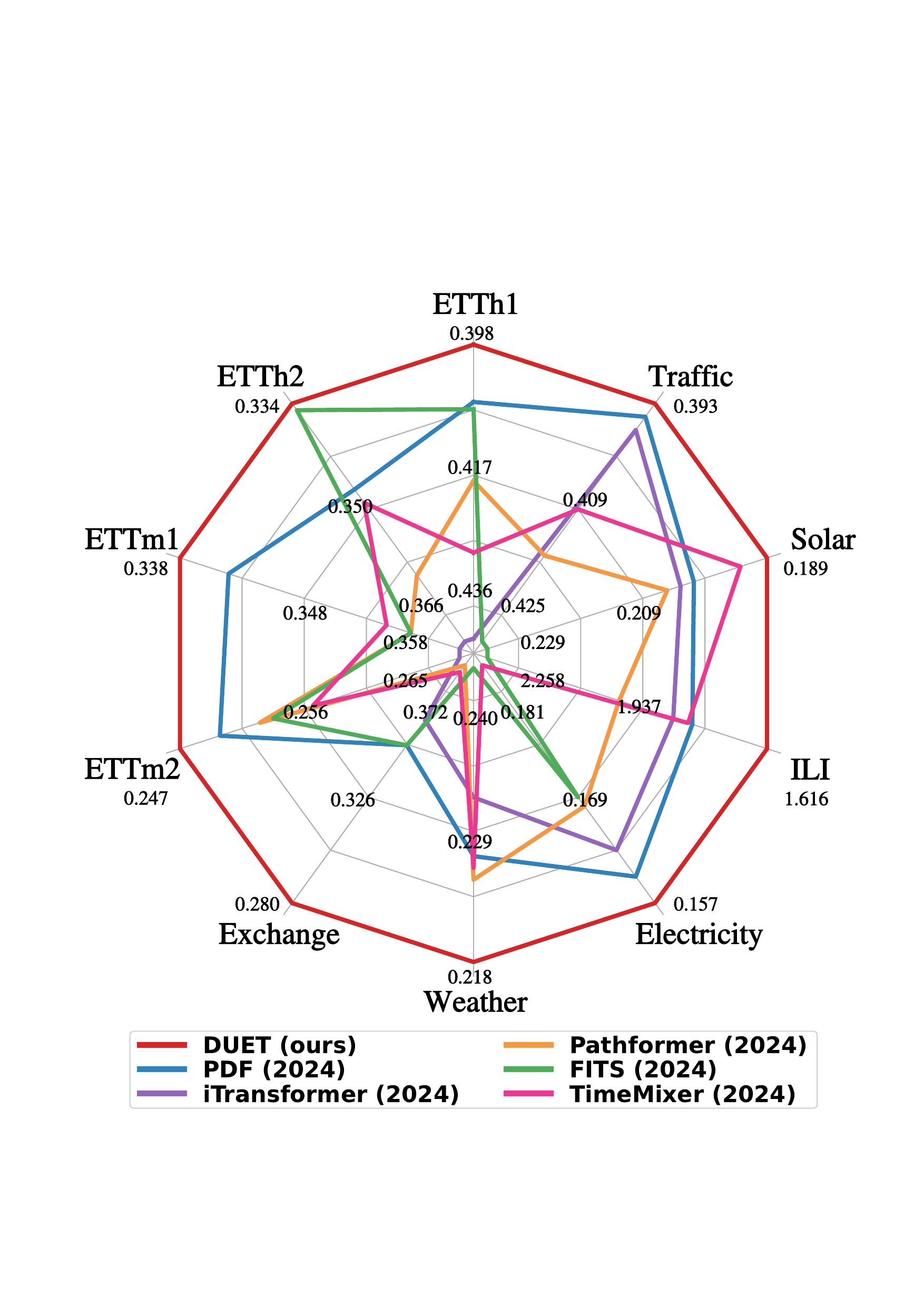}
  \caption{Performance of DUET. Results (MSE) are averaged from all forecasting horizons. DUET outperforms strong baselines in 10 commonly used datasets.}
  \label{zhizhu}
  \vspace{-6mm}
\end{figure}

In this study, we address the above two challenges by proposing a general framework, \textbf{DUET}, which introduces \textit{\underline{DU}al clustering on the temporal and channel dimensions to \underline{E}nhance multivariate \underline{T}ime series forecasting}. 
First, to model heterogeneous temporal patterns caused by TDS, we design a \textit{Temporal Clustering Module}~(TCM). This module clusters time series into fine-grained distributions, allowing us to use various pattern extractors to capture their intrinsic temporal patterns, thereby modeling the heterogeneity of temporal patterns. This method effectively handles both stationary and non-stationary data, demonstrating strong generality.
Second, to flexibly model relationships among channels, we propose a \textit{Channel Clustering Module}~(CCM). Using a channel-soft-clustering strategy, this module captures relationships among channels in the frequency domain through metric learning and applies sparsification. This approach enables each channel to focus on those beneficial for downstream prediction tasks, while mitigating the impact of noisy or irrelevant channels, thereby achieving effective channel soft clustering. 
Finally, the \textit{Fusion Module}~(FM), based on a masked attention mechanism, efficiently combines the temporal features extracted by the TCM with the channel mask matrix generated by the CCM.
Experimental results show that the proposed DUET achieves SOTA performance on real-world forecasting datasets---see Figure~\ref{zhizhu}.

Our contributions are summarized as follows. 
\begin{itemize}[left=0.1cm]
\item To address MTSF, we propose a general framework called DUET. It learns an accurate and adaptive forecasting model through dual clustering on both temporal and channel dimensions.
\item We design the TCM that clusters time series into fine-grained distributions. Various pattern extractors are then designed for different distribution clusters to capture their unique temporal patterns, modeling the heterogeneity of temporal patterns.

\item We design the CCM that flexibly captures the relationships among channels in the frequency domain through metric learning and applies sparsification.

\item We conduct extensive experiments on 25 datasets. The results show that DUET outperforms state-of-the-art baselines. Additionally, all datasets and code are avaliable at \url{https://github.com/decisionintelligence/DUET}.
\end{itemize}

\section{Related Works}
\subsection{Temporal Distribution Shift in MTSF}
Time series forecasting suffers from \textit{Temporal Distribution Shift} (TDS), as the distribution of real-world series changes over time~\cite{akay2007grey, dai2024ddn, TimeBridge}. In recent years, various methods have been proposed to address this issue.
Some works tackle TDS from a \textbf{normalization perspective}. DAIN~\cite{passalis2019deep} adaptively normalizes the series with nonlinear neural networks. RevIN~\cite{kim2021reversible} utilizes instance normalization on input and output sequences by normalizing the input sequences and then denormalizing the model output sequences. Dish-TS~\cite{fan2023dish} identifies intra- and inter-space distribution shifts in time series and mitigates these issues by learning distribution coefficients. Non-stationary Transformer~\cite{liu2022non} presents de-stationary attention that incorporates non-stationary factors in self-attention, significantly improving transformer-based models.
Some works address TDS from a \textbf{distribution perspective}. DDG-DA~\cite{li2022ddg} predicts evolving data distribution in a domain adaptation fashion. AdaRNN~\cite{du2021adarnn} proposes an adaptive RNN to alleviate the impact of non-stationary factors by characterizing and matching distributions. Other works address TDS from a \textbf{time-varying model parameters perspective}. Triformer~\cite{cirstea2022triformer} proposes a light-weight approach to enable variable-specific model parameters, making it possible
to capture distinct temporal patterns from different variables.  ST-WA~\cite{cirstea2022towards} use distinct sets of model parameters for different time period.

Despite the effectiveness of existing methods, our work explicitly models heterogeneous temporal patterns separately under different distributions, which can further improve the performance.

\subsection{Channel Strategies in MTSF}
It is essential to consider the correlations among channels in MTSF. Most existing methods adopt either a \textit{Channel-Independent} (\textbf{CI}) or \textit{Channel-Dependent} (\textbf{CD})  strategy to utilize the spectrum of information in channels. CI strategy approaches~\cite{xu2024fitsmodelingtimeseries,nie2022time,lin2024sparsetsf,PDFliu} share the same weights across all channels and make forecasts independently. Conversely, CD strategy approaches~\cite{chen2024pathformer,cirstea2022triformer,zhang2022crossformer,hu2024adaptive,liuicassp,liu2023itransformer} consider all channels simultaneously and generates joint representations for decoding. The CI strategy is characterized by low model capacity but high robustness, whereas the CD strategy exhibits the opposite characteristics. DGCformer~\cite{liu2024dgcformer} proposes relatively balanced channel strategies called \textit{Channel-Hard-Clustering} (\textbf{CHC}), trying to mitigate this polarization effect and improve predictive capabilities. Specifically, DGCformer designs a graph clustering module to assign channels with significant similarities into the same cluster, utilizing the CD strategy inside each cluster and the CI strategy across them. This approach adopts the CHC strategy, focusing solely on channel correlations within the same cluster. As a result, this method suffers from the limitations of rigidly adhering to channel-similarity rules defined by human experience.

Different from the above methods, we adopt a Channel-Soft-Clustering (CSC) strategy and devise a fully adaptive sparsity module to dynamically build group for each channel, which is a more comprehensive design covering the CHC strategy.

\begin{figure*}[t!]
  \centering
  \includegraphics[width=0.95\linewidth]{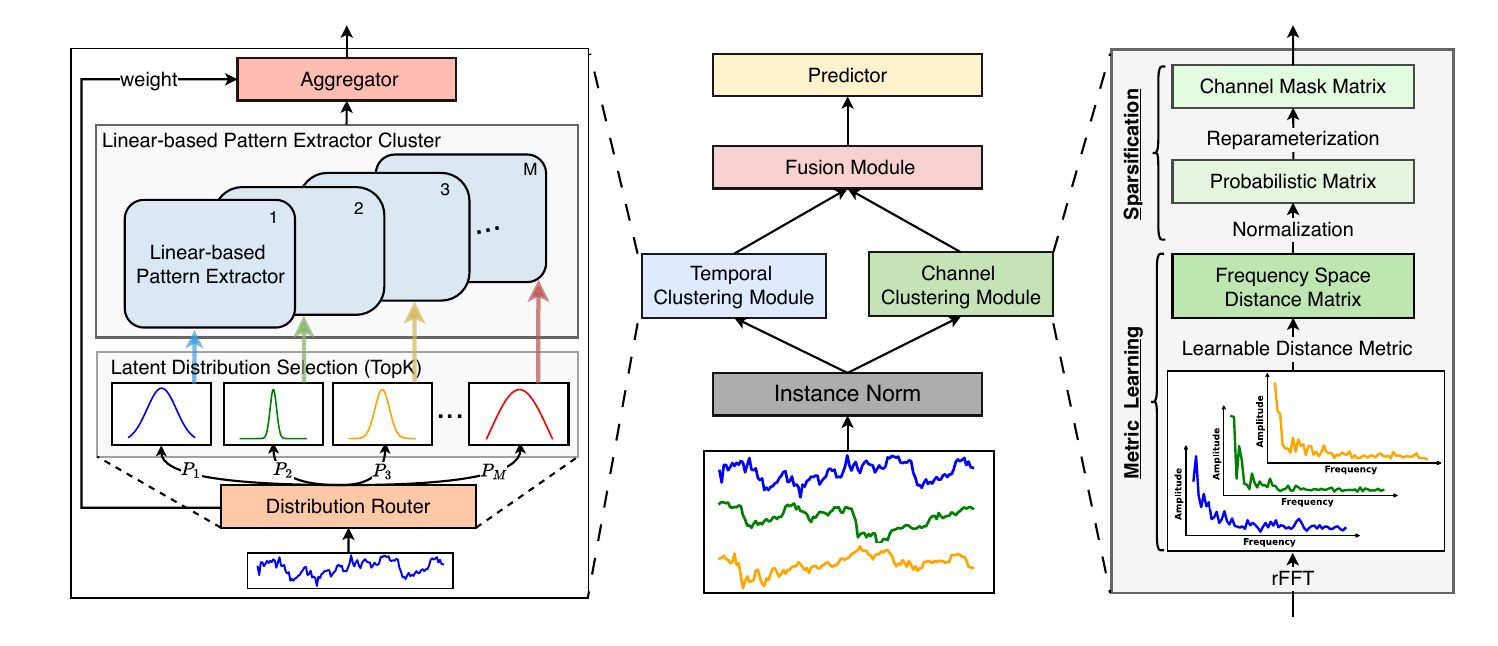}
  \caption{The architecture of DUET. Temporal Clustering Module clusters time series into fine-grained distribution. For different distribution clusters, various pattern extractors are designed to capture their intrinsic temporal patterns. Channel Clustering Module flexibly captures the relationships among channels in the frequency domain space through Metric Learning and applies Sparsification. Fusion Module combines the temporal features and the channel mask matrix.}
    \label{fig:overview}
\end{figure*}

\section{Preliminaries}
\subsection{Definitions}
\begin{definition}[Time series]
A time series  $X \in \mathbb{R}^{N\times T}$ is a time-oriented sequence of N-dimensional time points,  where $T$ is the number of timestamps, and $N$ is the number of channels. If $N=1$, a time series is called univariate, and multivariate if $N>1$. 
\end{definition}
For convenience, we separate dimensions with commas. Specifically, we denote $X_{i,j} \in \mathbb{R}$ as the $i$-th channel at the $j$-th timestamp, $X_{n,:}\in \mathbb{R}^T$ as the time series of $n$-th channel, where $n=1,\cdots,N$. We also introduce some definitions used in our methodology:
\begin{definition}[Temporal Distribution Shift~\cite{du2021adarnn}]
\label{def: temporal distribution shift}
Given a time series $\mathcal{X} \in \mathbb{R}^{N \times L}$, by sliding the window, we get a set of time series with the length of $T$, denoted as $\mathcal{D} = \{\mathcal{X}_{n,i:i+T} | n\in [1, N] \& i\in [1,L-T]  \}$, where each $\mathcal{X}_{n,i:i+T}$ equals to such $X_{n,:}$. Then, temporal distribution shift is referred to the case that $\mathcal{D}$ can be clustered into $K$ sets, i.e., $\mathcal{D}=\bigcup_{i=1}^K \mathcal{D}_i$, where each $\mathcal{D}_i$ denotes the set with data distribution $P_{\mathcal{D}_i}(x)$, where $P_{\mathcal{D}_i}(x) \ne P_{\mathcal{D}_j}(x), \forall i \ne j$ and $1\leq i,j \leq k$. 
\end{definition}

\subsection{Problem Statement}
\textbf{Multivariate Time Series Forecasting} aims to predict the next $F$ future timestamps, formulated as $Y =\langle X_{:,T+1}, \cdots, X_{:,T+F}\rangle~\in \mathbb{R}^{N \times F}$ based on the historical time series $X =~\langle X_{:,1}, \cdots, X_{:,T}\rangle~\in \mathbb{R}^{N \times T}$ with $N$ channels and $T$ timestamps.

\section{Methodology}
\subsection{Structure Overview}
Figure~\ref{fig:overview} shows the architecture of DUET, which adopts a dual clustering on both temporal and channel dimensions, simultaneously mining intrinsic temporal patterns and dynamic channel correlations. Specifically, we first use the Instance Norm~\cite{kim2021reversible} to unify the distribution of training and testing data. Then, the Temporal Clustering Module (TCM) utilizes a specially designed Distribution Router (Figure~\ref{Distribution Router}) to capture the potential latent distributions of each time series $X_{n,:}\in\mathbb{R}^T$ in a channel-independent way, and then clusters time series with similar latent distributions by assigning them to the same group of Linear-based Pattern Extractors (Figure~\ref{Linear-based Pattern Extractor}). In this way, we can mitigate the issue that single structure cannot fully extract temporal features due to heterogeneity of temporal patterns, even with millions of parameters. Meanwhile, the Channel Clustering Module (CCM) captures the correlations among channels in the frequency space in a channel soft clustering way. By leveraging an adaptive metric learning technique and applying sparsification, the CCM outputs a learned channel-mask matrix, so each channel can focus on those beneficial for downstream prediction and isolate the adverse effects of irrelevant channels in a sparse connection way. Finally, the Fusion Module (FM), based on a masked attention mechanism (Figure~\ref{temporal-channel Fusion}), effectively combines the temporal features extracted by the TCM and the channel-mask matrix generated by the CCM. A linear predictor is then used to forecast the future values at the end of our framework.

According to the above-mentioned description, the process of DUET can be formulated as follows:
\begin{align}
  X^{\textit{norm}} &= InstanceNorm(X) \label{1}\\
 X^{\textit{temp}} &= TCM(X^{\textit{norm}}) \label{2}\\
\mathcal{M} &= CCM(X^{\textit{norm}})\label{3} \\
X^{\textit{mix}} = FM(&X^{\textit{temp}},\ \mathcal{M}), \ \hat{Y} = Predictor(X^{\textit{mix}}),  
\end{align}
where $X^{\textit{temp}}, X^{\textit{mix}}\in \mathbb{R}^{N \times d}, \mathcal{M}\in \mathbb{R}^{N \times N}$. Steps \ref{2} and \ref{3} can be computed simultaneously for higher efficiency. Below, we will introduce the details of each module in our framework.

\subsection{Temporal Clustering Module (TCM)}
\begin{figure*}[t!]
  \centering
  \subfloat[Distribution Router]
  {\includegraphics[width=0.3\textwidth,height=0.32\linewidth]{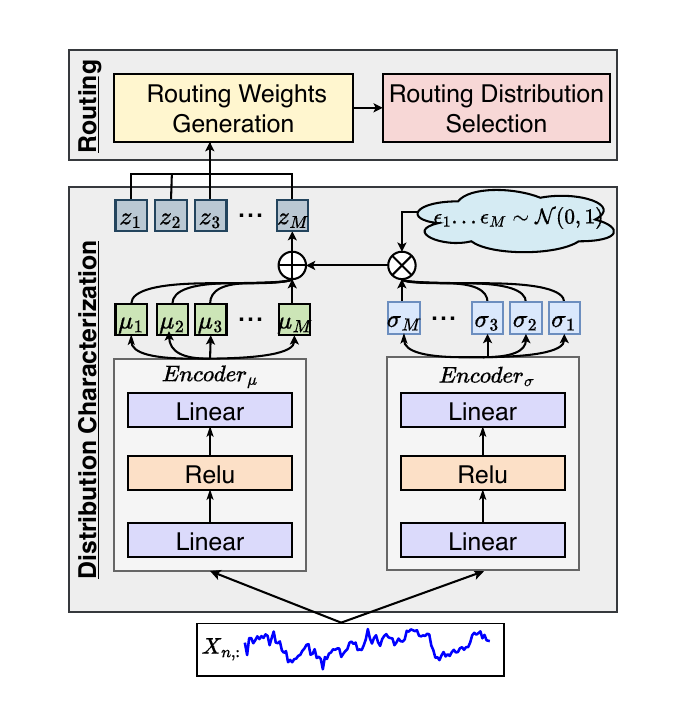}  \label{Distribution Router}}
  \subfloat[Linear Pattern Extractor]
  {\includegraphics[width=0.23\textwidth,height=0.33\linewidth]{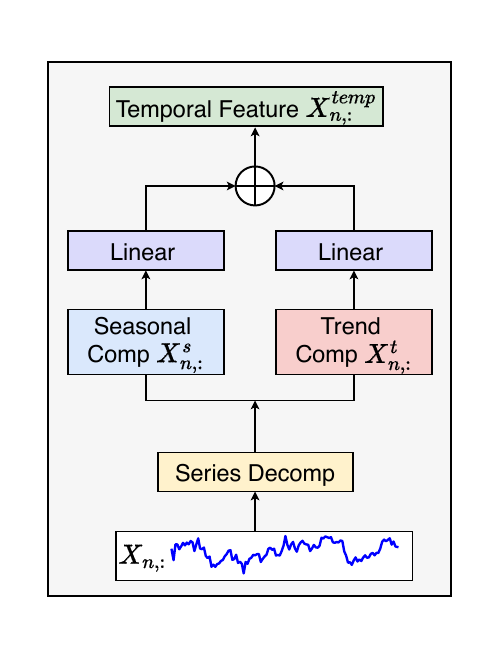}\label{Linear-based Pattern Extractor}}
  \subfloat[Learnable Distance Metric]
  {\includegraphics[width=0.23\textwidth,height=0.33\linewidth]{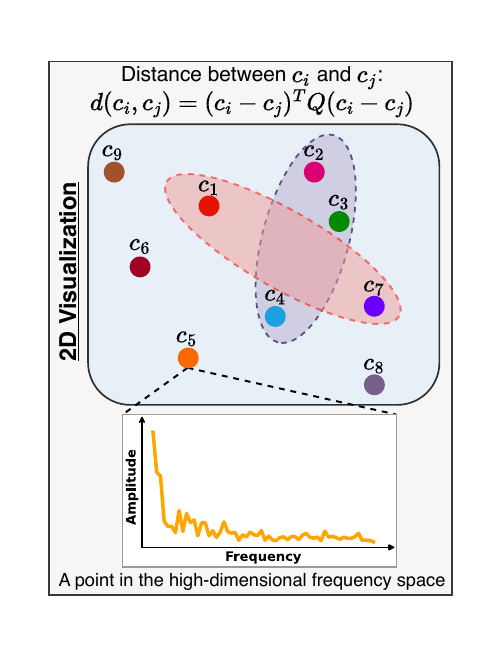}\label{Learnable distance metric}}
  \subfloat[Fusion Module]
  {\includegraphics[width=0.235\textwidth,height=0.33\linewidth]{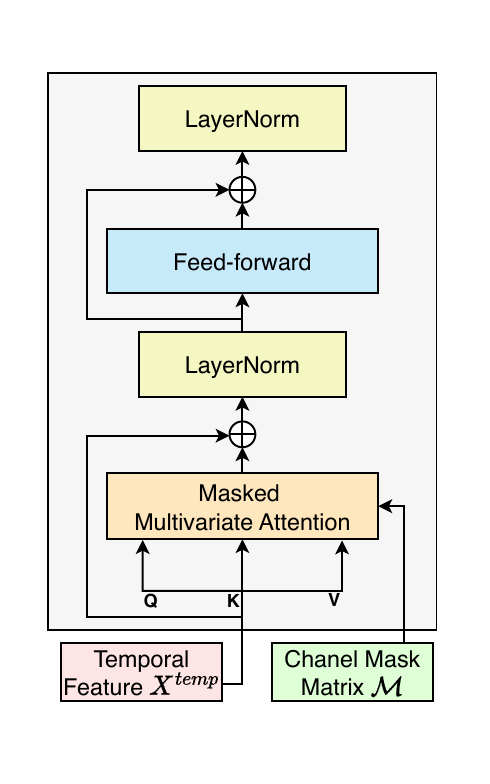}\label{temporal-channel Fusion}}
\caption{(a) The structure of the Distribution Router, which consists of Distribution Characterization and Routing. (b) The structure of the Linear-based Pattern Extractor, which decomposes the series into seasonal and trend parts, separately extracts temporal features with linear models and reads them. (c) The Learnable Distance Metric is to capture the relationships among channels. (d) The Fusion Module is to combine the temporal features and the channel mask matrix.}
\label{Example of dataset label revise.}
\end{figure*}

To model heterogeneous temporal patterns
caused by TDS, we design a Linear-based Pattern Extractor Cluster, where each Linear-based Pattern Extractor extracts temporal features for time series that has the same latent distribution. We adopt a linear model as the basic structure because it has been proven to efficiently extract temporal information~\cite{wang2024timemixer, xu2024fitsmodelingtimeseries, lin2024sparsetsf, zeng2023transformers,lincyclenet}, which also helps to keep the cluster lightweight. Moreover, we design a simple yet effective Distribution Router---see Figure~\ref{Distribution Router}, which can extract the potential latent distributions of the current time series and determine the corresponding Linear-based Pattern Extractor. Both the routing and extracting processes are conducted in a channel-independent way, focusing on clustering univariate time series $X_{n,:}\in \mathbb{R}^{T}$ based on their distributions and fully extracting their temporal patterns.

\noindent
\textbf{Distribution Router:} Inspired by VAE~\cite{kingma2013auto}, we first design two fully connected layer-based encoders to adaptively capture $M$ candidate latent distributions for each time series \(X_{n,:} \in \mathbb{R}^{T}\) (see Figure~\ref{Distribution Router}), where \(M\) denotes the size of the Linear-based Pattern Extractor Cluster. Empirically, we assume each time series follows a latent normal distribution and the process can be formulated as follows. 
\begin{align}
    \textit{Encoder}_{\mu}(X_{n,:}) &= \textit{ReLU}(X_{n,:}\cdot W_0^\mu)\cdot W_1^{\mu}\\
    \textit{Encoder}_{\sigma}(X_{n,:}) &= \textit{ReLU}(X_{n,:}\cdot W_0^\sigma)\cdot W_1^{\sigma},
\end{align}
where $W_0^\mu,W_0^\sigma \in \mathbb{R}^{T\times d_0}$, $W_1^\mu,W_1^\sigma \in \mathbb{R}^{d_0\times M}$. Then, we utilize the Noisy Gating technique~\cite{shazeer2017outrageously} to choose $k$ most possible distributions which \(X_{n,:}\) potentially belongs to and calculate the corresponding weights. We can observe that the reparameterization trick used to measure normal distributions shares similarities with the noise addition technique in Noisy Gating, so we elegantly combine them in a unified form:
\begin{align}    
Z_n &=\textit{Encoder}_{\mu}(X_{n,:}) + \epsilon \odot \textit{Softplus}(\textit{Encoder}_{\sigma}(X_{n,:}))\\
H(X_{n,:}) &= W^H \cdot Z_n,
\end{align}
where \(H(X_{n,:}),\epsilon \in \mathbb{R}^M, \epsilon_i \sim \mathcal{N}(0,1)\), $W^H\in \mathbb{R}^{M \times M}$. The introduction of \(\epsilon\) not only facilitates resampling from a normal distribution but also stabilizes the training process of Noisy Gating. The activation function $\textit{Softplus}$ helps keep the variance to be positive. \(H(X_{n,:})\) denotes the projected weights of distributions. We demonstrate the computational equivalence to the original Noisy Gating~\cite{shazeer2017outrageously} in Appendix~\ref{app:noisygating}. Subsequently, we select $k$ ($k \leq M$) most possible latent distributions and calculate the weights:
\begin{align}
G(X_{n,:}) &= \textit{Softmax}(\textit{KeepTopK}(H(X_{n,:}),k))\\
\textit{KeepTopK}(H(X_{n,:}),k)_i &= \begin{cases}
H(X_{n,:})_i & \text{if } i \in \textit{ArgTopk}(H(X_{n,:})) \\
-\infty & \text{otherwise}
\end{cases},
\end{align}
where $G(X_{n,:}) \in \mathbb{R}^{k}$ denotes the probabilities of each candidate distributions.
From the perspective of clustering, univariate time series $X_{n,:}\in \mathbb{R}^T,n=1,\cdots,N$ belonging to the same $k$ most possible latent distributions tend to be processed by the same group of $k$ Linear-based Pattern Extractors.

\noindent
\textbf{Linear-based Pattern Extractor:} The above-mentioned distribution router determines the latent distributions. Then, \(X_{n,:}\) is passed to corresponding $k$ selected Linear-based Pattern Extractors for temporal feature extraction. For the $i$-th Linear-based Pattern Extractor---see Figure~\ref{Linear-based Pattern Extractor}, we first use the moving average technique to decompose \(X_{n,:}\) into seasonal and trend parts~\cite{wu2021autoformer, zeng2023transformers}, then separately extract features and finally fuse them to better capture the patterns of time series from the same distribution:
\begin{align}
X^t_{n,:} &= \textit{AvgPool}(\textit{padding}(X_{n,:}))\\
X^s_{n,:} &= X_{n,:} - X^t_{n,:}\\
X^{\textit{temp}^i}_{n,:} &= X_{n,:}^t \cdot W_t^i + X_{n,:}^s\cdot W_s^i,
\end{align}
where $X^t_{n,:},X^s_{n,:} \in \mathbb{R}^{T}$ are the decomposed trend and seasonal parts, $W_t^i,W_s^i \in \mathbb{R}^{T\times d}$ are the learnable parameters in linear transformations. $X^{\textit{temp}^i}_{n,:} \in \mathbb{R}^{d}$ is the temporal feature generated by the $i$-th extractor, where $d$ is the hidden dimension. 

\noindent
\textbf{Aggregator:} After the selected $k$ Linear-based Pattern Extractors output the corresponding features, the Aggregator gathers the temporal features based on the previously calculated weighted gates:
\begin{align}
X^{\textit{temp}}_{n,:}  = \sum_{i=1}^k G(X_{n,:})_i \cdot X^{\textit{temp}^i}_{n,:},
\end{align}
where $G(X_{n,:})_i$ represents the weights of the \(i\)-th Linear-based Pattern Extractor, and $X^{\textit{temp}}_{n,:} \in \mathbb{R}^{d}$ represents the temporal patterns of $X_{n,:}$ extracted by the Linear-based Pattern Extractor Cluster. After obtaining temporal features of each $X_{n,:}$ ($n=1,\cdots,N$) in a channel-independent way, they are finally gathered into $X^{\textit{temp}} \in \mathbb{R}^{N\times d}$.

\subsection{Channel Clustering Module (CCM)}
To mitigate the adverse effects caused by improper consideration of cross-channel relationships during prediction, we devise an efficient metric learning method to softly cluster channels in the frequency space, and generate a corresponding learned channel-mask matrix to achieve sparse connections.\par
\noindent
\textbf{Learnable Distance Metric}. 
To fully utilize the cross-channel enhancement in prediction, we manage to leverage more useful information by modeling the correlations among channels in the perspective of frequency metric space. Theoretically, the ``frequency space'' describes functions under the fourier basis, and their coordinates come from Fourier Transform by computing integrals in $L^2$ space (see Appendix~\ref{app:fourier}). Specifically, given a discrete time series at the $n$-th channel, $X_{n,:} \in \mathbb{R}^{T}$, we first use real Fast Fourier Transform~(rFFT~\cite{rfft}) to project it into a finite $\frac{T}{2}$-dimension frequency space with the same $\frac{T}{2}$ Fourier basis (in complex form). Then, we need to find a proper distance metric that precisely evaluates channel relations in the frequency space.\par

We propose such a distance metric that precisely evaluates channel relations in the frequency space and make each channel obtain maximum neighbour gain in the prediction task. Specifically, we take the norm (or amplitude) to obtain \( X^{\textit{chan}}_{n,:} \in \mathbb{R}^{\frac{T}{2}} \) and adopt a learnable Mahalanobis distance metric~\cite{goldberger2004neighbourhood} to adaptively discover the interrelationships among channels: 
\begin{align}
     d(X_{i,:},X_{j,:}) &= (X^{\textit{chan}}_{i,:}-X^{\textit{chan}}_{j,:})^T \cdot Q \cdot (X^{\textit{chan}}_{i,:}-X^{\textit{chan}}_{j,:})\\ 
     X^{\textit{chan}}_{i,:} &= \textit{norm}(\textit{rFFT}(X_{i,:})), \ 
 X^{\textit{chan}}_{j,:}=\textit{norm}(\textit{rFFT}(X_{j,:})),
\end{align}
where $Q \in \mathbb{R}^{\frac{T}{2} \times \frac{T}{2}}$ is a learnable semi-positive definite matrix. It can be practically constructed by $Q = A^T\cdot A$, where $A$ is also a learnable matrix. This process introduces a more general and lightweight method from the perspective of metric space and adaptively explores a distance metric to measure the channel correlations for better prediction accuracy. \par
\noindent
\textbf{Normalization.} With the learnable channel distance metric, we first compute the relationship matrix of channels and normalize it to the range of [0, 1]:
\begin{align}
     D_{ij} &= d(X_{i,:},X_{j,:}) \\
 C_{ij} &= \begin{cases}
\frac{1}{D_{ij}} & \text{if } i \ne j \\
0 & \text{if } i = j
\end{cases},\ P_{ij} = \begin{cases}
\frac{C_{ij}\cdot \gamma}{\max_j(C_{ij})} & \text{if } i \ne j \\
1 & \text{if } i = j
\end{cases},
\end{align}
where $D,C,P \in \mathbb{R}^{N \times N}$ are distance, relationship, and probability matrices, $\gamma \in (0,1)$ is a discount factor to avoid the absolute connection. Above processes probabilize the relationships among channels, where \( P_{ij} \) represents the probability that channel $j$ is useful for channel $i$ in the prediction task. \par

\noindent
\textbf{Reparameterization.} Since our goal is to filter out the adverse effects of irrelevant channels and retain the beneficial effects of relevant channels, we further perform Bernoulli resampling on the probability matrix to obtain a binary channel mask matrix $\mathcal{M}\in \mathbb{R}^{N \times N}$, where \( \mathcal{M}_{ij} \approx \text{Bernoulli}(P_{ij}) \). Higher probability \( P_{ij} \) results in \( \mathcal{M}_{ij} \) closer to 1, indicating a relationship between channel $i$ and channel $j$. Since \( P_{ij} \) contains learnable parameters, we use the Gumbel Softmax reparameterization trick~\cite{jang2016categorical} during Bernoulli resampling to ensure the propagation of gradients. 

\subsection{Fusion Module (FM)}
Through the dual clustering on both temporal and channel dimensions, our framework extracts the temporal feature \( X^{\textit{temp}} \in \mathbb{R}^{N \times d} \) and a channel mask matrix \( \mathcal{M} \in \mathbb{R}^{N \times N} \) for each time series $X \in \mathbb{R}^{N\times T}$. Subsequently, we utilize a masked attention mechanism to further fuse them. This process can be formalized as follows.
\begin{align}
Q = X^{\textit{temp}} \cdot W^Q,\  K &= X^{\textit{temp}} \cdot W^K,\  V = X^{\textit{temp}} \cdot W^V \\
 \textit{MaskedScores} &= \frac{Q\cdot K^T}{\sqrt{d}} \odot \mathcal{M} + (1 - \mathcal{M})\odot (-\infty) \\ 
 X^{\textit{mix}} &= \textit{Softmax(MaskedScores)} \cdot V,
\end{align}
where $W^Q, W^K, W^V \in \mathbb{R}^{d \times d}$ are projection matrices in the attention block, $\textit{MaskedScores} \in \mathbb{R}^{N \times N}$ is the attention score matrix, $X^{\textit{mix}} \in \mathbb{R}^{N \times d}$ is the fused feature. With the application of the masked attention mechanism~\cite{vaswani2017attention}, the Fusion Module effectively fuses the temporal features extracted by TCM based on the sparse correlations captured by CCM. We also adopt LayerNorm, Feed-Forward and skip-connection in the Fusion Module (FM) like a classic transformer block. Specifically, LayerNorm~\cite{ba2016layer} and Skip-Connection~\cite{zagoruyko2016wide} ensure FM's stability and robustness, while the Feed-Forward layer~\cite{das2023long} empowers the FM to capture complex features. Their synergistic operation optimizes the representational capacity and training efficiency of the Fusion Module. \par
Finally, we adopt a linear projection to predict the future values, which is formulated as follows.
\begin{align}
     \hat{Y} &= X^{\textit{mix}} \cdot  W^O,
\end{align}
where $W^O \in \mathbb{R}^{d \times F}$, $\hat{Y}\in \mathbb{R}^{N\times F}$.

\renewcommand{\arraystretch}{1} %
\begin{table*}[t!]
\caption{Statistics of multivariate datasets.}
\label{Multivariate datasets}
\resizebox{1.84\columnwidth}{!}{
\begin{tabular}{@{}lllrrcl@{}}
\toprule
Dataset      & Domain      & Frequency & Lengths & Dim & Split  & Description\\ \midrule
METR-LA      & Traffic     & 5 mins    & 34,272      & 207      & 7:1:2  & Traffic speed dataset collected from loopdetectors in the LA County road network\\
PEMS-BAY     & Traffic     & 5 mins    & 52,116      & 325      & 7:1:2 &Traffic speed dataset collected from the CalTrans PeMS\\
PEMS04       & Traffic     & 5 mins    & 16,992      & 307      & 6:2:2 &Traffic
flow time series collected from the CalTrans PeMS\\
PEMS08       & Traffic     & 5 mins    & 17,856      & 170      & 6:2:2 &Traffic
flow time series collected from the CalTrans PeMS\\
Traffic      & Traffic     & 1 hour    & 17,544      & 862      & 7:1:2 & Road occupancy rates measured by 862 sensors on San Francisco Bay area freeways\\
ETTh1        & Electricity & 1 hour     & 14,400      & 7        & 6:2:2 & Power transformer 1, comprising seven indicators such as oil temperature and useful load\\
ETTh2        & Electricity & 1 hour    & 14,400      & 7        & 6:2:2 & Power transformer 2, comprising seven indicators such as oil temperature and useful load\\
ETTm1        & Electricity & 15 mins   & 57,600      & 7        & 6:2:2 & Power transformer 1, comprising seven indicators such as oil temperature and useful load\\
ETTm2        & Electricity & 15 mins   & 57,600      & 7        & 6:2:2 & Power transformer 2, comprising seven indicators such as oil temperature and useful load\\
Electricity  & Electricity & 1 hour    & 26,304      & 321      & 7:1:2 & Electricity records the electricity consumption in kWh every 1 hour from 2012 to 2014\\
Solar        & Energy      & 10 mins   & 52,560      & 137      & 6:2:2 &Solar production records collected from 137 PV plants in Alabama \\
Wind         & Energy & 15 mins   & 48,673      & 7        & 7:1:2 & Wind power records from 2020-2021 at 15-minute intervals \\
Weather      & Environment & 10 mins   & 52,696      & 21       & 7:1:2 & Recorded
every for the whole year 2020, which contains 21 meteorological indicators\\
AQShunyi     & Environment & 1 hour    & 35,064      & 11       & 6:2:2 & Air quality datasets
from a  measurement station, over a period of 4 years\\
AQWan       & Environment & 1 hour    & 35,064      & 11       & 6:2:2 & Air quality datasets
from a  measurement station, over a period of 4 years\\
ZafNoo      & Nature      & 30 mins   & 19,225      & 11       & 7:1:2 & From the Sapflux
data project includes sap flow measurements and nvironmental variables\\
CzeLan       & Nature      & 30 mins   & 19,934      & 11       & 7:1:2 & From the Sapflux
data project includes sap flow measurements and nvironmental variables\\
FRED-MD      & Economic    & 1 month   & 728        & 107      & 7:1:2 & Time series showing a set of macroeconomic indicators from the Federal Reserve Bank\\
Exchange & Economic    & 1 day      & 7,588       & 8        & 7:1:2 & ExchangeRate collects the daily exchange rates of eight countries\\
NASDAQ      & Stock       & 1 day     & 1,244       & 5        & 7:1:2 & Records opening price, closing price, trading volume, lowest price, and highest price\\
NYSE         & Stock       & 1 day     & 1,243       & 5        & 7:1:2 & Records opening price, closing price, trading volume, lowest price, and highest price\\
NN5         & Banking     & 1 day     & 791        & 111      & 7:1:2 & NN5 is from banking, records the daily cash withdrawals from ATMs in UK\\
ILI          & Health      & 1 week     & 966        & 7        & 7:1:2 & Recorded indicators of patients data from Centers for Disease Control and Prevention\\
Covid-19     & Health      & 1 day     & 1,392       & 948      & 7:1:2 & Provide opportunities for researchers to investigate the dynamics of COVID-19\\ 
Wike2000    & Web         & 1 day     & 792        & 2,000     & 7:1:2 & Wike2000 is daily page views of 2000 Wikipedia pages \\ \bottomrule
\end{tabular}}
\end{table*}

\renewcommand{\arraystretch}{1}
\begin{table}[t!]
\caption{Ablation study. Results are averaged from all forecasting horizons. }
\resizebox{1\columnwidth}{!}{
\footnotesize
\begin{tabular}{cc|cccccccccc}
\toprule
  \multicolumn{2}{c|}{\textbf{Model}} & \multicolumn{2}{c}{\textbf{DUET}} & \multicolumn{2}{c}{\textbf{w/o  TCM}} & \multicolumn{2}{c}{\textbf{w/o  CCM}} & \multicolumn{2}{c}{\textbf{Full Attention}} & \multicolumn{2}{c}{\textbf{Tem Info}}  \\
\multicolumn{2}{c|}{Metrics}  & mse & mae & mse & mae & mse & mae & mse & mae & mse & mae   \\

\midrule
\multicolumn{2}{c|}{ETTh2}&  \textbf{0.334} &	\textbf{0.383} 	&0.344 &	0.391 &	0.343 &	0.391 &	0.344 	&0.389 &	0.345 &	0.390  \\
 \addlinespace\cline{1-12} \addlinespace
\multicolumn{2}{c|}{ETTm2}& \textbf{0.247}  &	\textbf{0.307}  &	0.256  &	0.310  &	0.256  &	0.312  &	0.254  &	0.310 	 &0.254 	 &0.311  \\
 \addlinespace\cline{1-12} \addlinespace
\multicolumn{2}{c|}{Weather}&\textbf{0.218} &	\textbf{0.252} &	0.225 &	0.255 &	0.232 &	0.262 &	0.223 &	0.255 &	0.222 &	0.254 \\
 \addlinespace\cline{1-12} \addlinespace 
\multicolumn{2}{c|}{Traffic}&  \textbf{0.393} &	\textbf{0.256} &	0.398 &	0.263 &	0.439 &	0.277 &	0.396 &	0.260 &	0.401 &	0.263 \\
\addlinespace
\bottomrule
\end{tabular}
}
\label{tab: Ablation study average results}
\end{table}

\section{Experiments}
\subsection{Experimental Settings}
\subsubsection{Datasets}
We use 25 real-world datasets from 10 different domains in time series forecasting benchmark (TFB)~\cite{qiu2024tfb} to comprehensively evaluate the performance of DUET, more details of the benchmark datasets are included in Table~\ref{Multivariate datasets}.
Due to space constraints, we report the results on 10 well-acknowledged forecasting datasets which include ETT (ETTh1, ETTh2, ETTm1, ETTm2), Exchange, Weather, Electricity, ILI, Traffic, and Solar in the main text. 
The results for the remaining 15 datasets are available in our code repository at ~\url{https://github.com/decisionintelligence/DUET}.
\subsubsection{Baselines}
We choose the latest state-of-the-art models to serve as baselines, including CNN-based models~(TimesNet~\cite{wu2022timesnet}), MLP-based models (FITS~\cite{xu2024fitsmodelingtimeseries}, TimeMixer~\cite{wang2024timemixer}, and DLinear~\cite{zeng2023transformers}), and Transformer-based models~(PDF~\cite{PDFliu}, iTransformer~\cite{zhou2021informer}, Pathformer~\cite{chen2024pathformer}, PatchTST~\cite{nie2022time}, Crossformer~\cite{zhang2022crossformer}, and Non-stationary Transformer (Stationary)~\cite{liu2022non}).
\subsubsection{Implementation Details}
To keep consistent with previous works, we adopt Mean Squared Error (mse) and Mean Absolute Error (mae) as evaluation metrics. We consider four forecasting horizon $F$: {24, 36, 48, and 60} for FredMd, NASDAQ, NYSE, NN5, ILI, Covid-19, and Wike2000, and we use another four forecasting horizon, {96, 192, 336, and 720,} for all other datasets which have longer lengths. Since the size of the look-back window can affect the performance of different models, we choose the look-back window size in {36 and 104} for FredMd, NASDAQ, NYSE, NN5, ILI, Covid-19, and Wike2000, and {96, 336, and 512} for all other datasets and report each method's best results for fair comparisons. 

We utilize the TFB code repository for unified evaluation, with all baseline results also derived from TFB. Following the settings in TFB~\cite{qiu2024tfb} and FoundTS~\cite{li2024foundts}, we do not apply the ``Drop Last'' trick to ensure a fair comparison. All experiments of DUET are conducted using PyTorch~\cite{paszke2019pytorch} in Python 3.8 and executed on an NVIDIA Tesla-A800 GPU. The training process is guided by the L1 loss function and employs the ADAM optimizer. The initial batch size is set to 64, with the flexibility to halve it (down to a minimum of 8) in case of an Out-Of-Memory (OOM) issue.

\renewcommand{\arraystretch}{1}
\begin{table*}[h]
\caption{Multivariate forecasting results with forecasting horizons $F \in \{12, 24, 36, 48\}$ for ILI and $F \in \{96, 192, 336, 720\}$ for others. }
\label{Common Multivariate forecasting results avg}
\resizebox{2\columnwidth}{!}{
\begin{tabular}{cc|cc|cc|cc|cc|cc|cc|cc|cc|cc|cc|cc|}
\toprule
\multicolumn{2}{c|}{\multirow{2}{*}{Models}} & \multicolumn{2}{c}{DUET} & \multicolumn{2}{c}{PDF} & \multicolumn{2}{c}{iTransformer} & \multicolumn{2}{c}{Pathformer} & \multicolumn{2}{c}{FITS} & \multicolumn{2}{c}{TimeMixer} & \multicolumn{2}{c}{PatchTST} & \multicolumn{2}{c}{Crossformer} & \multicolumn{2}{c}{TimesNet} & \multicolumn{2}{c}{DLinear} & \multicolumn{2}{c}{Stationary} \\
\multicolumn{2}{c|}{} & \multicolumn{2}{c}{(ours)} & \multicolumn{2}{c}{(2024)} & \multicolumn{2}{c}{(2024)} & \multicolumn{2}{c}{(2024)} & \multicolumn{2}{c}{(2024)} & \multicolumn{2}{c}{(2024)} & \multicolumn{2}{c}{(2023)} & \multicolumn{2}{c}{(2023)} & \multicolumn{2}{c}{(2023)} & \multicolumn{2}{c}{(2023)} & \multicolumn{2}{c}{(2022)} \\
\addlinespace\cline{1-24} \addlinespace
\multicolumn{2}{c|}{Metrics} & mse & mae & mse & mae & mse & mae & mse & mae & mse & mae & mse & mae & mse & mae & mse & mae & mse & mae & mse & mae & mse & mae \\
\midrule

\multirow[c]{4}{*}{\rotatebox{90}{ETTh1}} & 96 & \textbf{0.352} & \textbf{0.384} & \underline{0.360} & \underline{0.391} & 0.386 & 0.405 & 0.372 & 0.392 & 0.376 & 0.396 & 0.372 & 0.401 & 0.377 & 0.397 & 0.411 & 0.435 & 0.389 & 0.412 & 0.379 & 0.403 & 0.591 & 0.524 \\ 
 & 192 & \underline{0.398} & \textbf{0.409} & \textbf{0.392} & \underline{0.414} & 0.424 & 0.440 & 0.408 & 0.415 & 0.400 & 0.418 & 0.413 & 0.430 & 0.409 & 0.425 & 0.409 & 0.438 & 0.440 & 0.443 & 0.408 & 0.419 & 0.615 & 0.540 \\ 
 & 336 & \textbf{0.414} & \textbf{0.426} & \underline{0.418} & 0.435 & 0.449 & 0.460 & 0.438 & \underline{0.434} & 0.419 & 0.435 & 0.438 & 0.450 & 0.431 & 0.444 & 0.433 & 0.457 & 0.523 & 0.487 & 0.440 & 0.440 & 0.632 & 0.551 \\ 
 & 720 & \textbf{0.429} & \textbf{0.455} & 0.456 & 0.462 & 0.495 & 0.487 & 0.450 & 0.463 & \underline{0.435} & \underline{0.458} & 0.486 & 0.484 & 0.457 & 0.477 & 0.501 & 0.514 & 0.521 & 0.495 & 0.471 & 0.493 & 0.828 & 0.658 \\ 
\addlinespace\cline{1-24} \addlinespace
\multirow[c]{4}{*}{\rotatebox{90}{ETTh2}} & 96 & \textbf{0.270} & \textbf{0.336} & 0.276 & 0.341 & 0.297 & 0.348 & 0.279 & \underline{0.336} & 0.277 & 0.345 & 0.281 & 0.351 & \underline{0.274} & 0.337 & 0.728 & 0.603 & 0.334 & 0.370 & 0.300 & 0.364 & 0.347 & 0.387 \\ 
 & 192 & \underline{0.332} & \textbf{0.374} & 0.339 & 0.382 & 0.372 & 0.403 & 0.345 & 0.380 & \textbf{0.331} & \underline{0.379} & 0.349 & 0.387 & 0.348 & 0.384 & 0.723 & 0.607 & 0.404 & 0.413 & 0.387 & 0.423 & 0.379 & 0.418 \\ 
 & 336 & \underline{0.353} & \underline{0.397} & 0.374 & 0.406 & 0.388 & 0.417 & 0.378 & 0.408 & \textbf{0.350} & \textbf{0.396} & 0.366 & 0.413 & 0.377 & 0.416 & 0.740 & 0.628 & 0.389 & 0.435 & 0.490 & 0.487 & 0.358 & 0.413 \\ 
 & 720 & \textbf{0.382} & \textbf{0.425} & 0.398 & 0.433 & 0.424 & 0.444 & 0.437 & 0.455 & \underline{0.382} & \underline{0.425} & 0.401 & 0.436 & 0.406 & 0.441 & 1.386 & 0.882 & 0.434 & 0.448 & 0.704 & 0.597 & 0.422 & 0.457 \\ 
\addlinespace\cline{1-24} \addlinespace
\multirow[c]{4}{*}{\rotatebox{90}{ETTm1}} & 96 & \textbf{0.279} & \textbf{0.333} & \underline{0.286} & 0.340 & 0.300 & 0.353 & 0.290 & \underline{0.335} & 0.303 & 0.345 & 0.293 & 0.345 & 0.289 & 0.343 & 0.314 & 0.367 & 0.340 & 0.378 & 0.300 & 0.345 & 0.415 & 0.410 \\ 
 & 192 & \textbf{0.320} & \textbf{0.358} & \underline{0.321} & 0.364 & 0.341 & 0.380 & 0.337 & \underline{0.363} & 0.337 & 0.365 & 0.335 & 0.372 & 0.329 & 0.368 & 0.374 & 0.410 & 0.392 & 0.404 & 0.336 & 0.366 & 0.494 & 0.451 \\ 
 & 336 & \textbf{0.348} & \textbf{0.377} & \underline{0.354} & \underline{0.383} & 0.374 & 0.396 & 0.374 & 0.384 & 0.368 & 0.384 & 0.368 & 0.386 & 0.362 & 0.390 & 0.413 & 0.432 & 0.423 & 0.426 & 0.367 & 0.386 & 0.577 & 0.490 \\ 
 & 720 & \textbf{0.405} & \textbf{0.408} & \underline{0.408} & 0.415 & 0.429 & 0.430 & 0.428 & 0.416 & 0.420 & \underline{0.413} & 0.426 & 0.417 & 0.416 & 0.423 & 0.753 & 0.613 & 0.475 & 0.453 & 0.419 & 0.416 & 0.636 & 0.535 \\ 
\addlinespace\cline{1-24} \addlinespace
\multirow[c]{4}{*}{\rotatebox{90}{ETTm2}} & 96 & \textbf{0.161} & \textbf{0.248} & \underline{0.163} & 0.251 & 0.175 & 0.266 & 0.164 & \underline{0.250} & 0.165 & 0.254 & 0.165 & 0.256 & 0.165 & 0.255 & 0.296 & 0.391 & 0.189 & 0.265 & 0.164 & 0.255 & 0.210 & 0.294 \\ 
 & 192 & \textbf{0.214} & \textbf{0.286} & \underline{0.219} & 0.290 & 0.242 & 0.312 & 0.219 & \underline{0.288} & 0.219 & 0.291 & 0.225 & 0.298 & 0.221 & 0.293 & 0.369 & 0.416 & 0.254 & 0.310 & 0.224 & 0.304 & 0.338 & 0.373 \\ 
 & 336 & \textbf{0.267} & \underline{0.321} & 0.269 & 0.330 & 0.282 & 0.337 & \underline{0.267} & \textbf{0.319} & 0.272 & 0.326 & 0.277 & 0.332 & 0.276 & 0.327 & 0.588 & 0.600 & 0.313 & 0.345 & 0.277 & 0.337 & 0.432 & 0.416 \\ 
 & 720 & \textbf{0.348} & \textbf{0.374} & \underline{0.349} & 0.382 & 0.375 & 0.394 & 0.361 & \underline{0.377} & 0.359 & 0.381 & 0.360 & 0.387 & 0.362 & 0.381 & 0.750 & 0.612 & 0.413 & 0.402 & 0.371 & 0.401 & 0.554 & 0.476 \\ 
\addlinespace\cline{1-24} \addlinespace
\multirow[c]{4}{*}{\rotatebox{90}{Exchange}} & 96 & \underline{0.080} & \textbf{0.198} & 0.083 & 0.200 & 0.086 & 0.205 & 0.088 & 0.208 & 0.082 & \underline{0.199} & 0.084 & 0.207 & \textbf{0.079} & 0.200 & 0.088 & 0.213 & 0.112 & 0.242 & 0.080 & 0.202 & 0.083 & 0.203 \\ 
 & 192 & 0.162 & \textbf{0.288} & 0.172 & 0.294 & 0.177 & 0.299 & 0.183 & 0.304 & 0.173 & 0.295 & 0.178 & 0.300 & \underline{0.159} & 0.289 & \textbf{0.157} & \underline{0.288} & 0.209 & 0.334 & 0.182 & 0.321 & 0.159 & 0.293 \\ 
 & 336 & \textbf{0.294} & \textbf{0.392} & 0.323 & 0.411 & 0.331 & 0.417 & 0.354 & 0.429 & 0.317 & 0.406 & 0.376 & 0.451 & \underline{0.297} & \underline{0.399} & 0.332 & 0.429 & 0.358 & 0.435 & 0.327 & 0.434 & 0.317 & 0.412 \\ 
 & 720 & \underline{0.583} & \textbf{0.580} & 0.820 & 0.682 & 0.846 & 0.693 & 0.909 & 0.716 & 0.825 & 0.684 & 0.884 & 0.707 & 0.751 & 0.650 & 0.980 & 0.762 & 0.944 & 0.736 & \textbf{0.578} & \underline{0.605} & 0.725 & 0.656 \\ 
\addlinespace\cline{1-24} \addlinespace
\multirow[c]{4}{*}{\rotatebox{90}{Weather}} & 96 & \underline{0.146} & \textbf{0.191} & 0.147 & 0.196 & 0.157 & 0.207 & 0.148 & \underline{0.195} & 0.172 & 0.225 & 0.147 & 0.198 & 0.149 & 0.196 & \textbf{0.143} & 0.210 & 0.168 & 0.214 & 0.170 & 0.230 & 0.188 & 0.242 \\ 
 & 192 & \textbf{0.188} & \textbf{0.231} & 0.193 & 0.240 & 0.200 & 0.248 & \underline{0.191} & \underline{0.235} & 0.215 & 0.261 & 0.192 & 0.243 & 0.191 & 0.239 & 0.198 & 0.260 & 0.219 & 0.262 & 0.216 & 0.273 & 0.241 & 0.290 \\ 
 & 336 & \textbf{0.234} & \textbf{0.268} & 0.245 & 0.280 & 0.252 & 0.287 & 0.243 & \underline{0.274} & 0.261 & 0.295 & 0.247 & 0.284 & \underline{0.242} & 0.279 & 0.258 & 0.314 & 0.278 & 0.302 & 0.258 & 0.307 & 0.341 & 0.341 \\ 
 & 720 & \textbf{0.305} & \textbf{0.319} & 0.323 & 0.334 & 0.320 & 0.336 & 0.318 & \underline{0.326} & 0.326 & 0.341 & 0.318 & 0.330 & \underline{0.312} & 0.330 & 0.335 & 0.385 & 0.353 & 0.351 & 0.323 & 0.362 & 0.403 & 0.388 \\ 
\addlinespace\cline{1-24} \addlinespace
\multirow[c]{4}{*}{\rotatebox{90}{Electricity}} & 96 & \textbf{0.128} & \textbf{0.219} & \underline{0.128} & \underline{0.222} & 0.134 & 0.230 & 0.135 & 0.222 & 0.139 & 0.237 & 0.153 & 0.256 & 0.143 & 0.247 & 0.134 & 0.231 & 0.169 & 0.271 & 0.140 & 0.237 & 0.171 & 0.274 \\ 
 & 192 & \textbf{0.145} & \textbf{0.235} & 0.147 & \underline{0.242} & 0.154 & 0.250 & 0.157 & 0.253 & 0.154 & 0.250 & 0.168 & 0.269 & 0.158 & 0.260 & \underline{0.146} & 0.243 & 0.180 & 0.280 & 0.154 & 0.251 & 0.180 & 0.283 \\ 
 & 336 & \textbf{0.163} & \textbf{0.255} & \underline{0.165} & \underline{0.260} & 0.169 & 0.265 & 0.170 & 0.267 & 0.170 & 0.268 & 0.189 & 0.291 & 0.168 & 0.267 & 0.165 & 0.264 & 0.204 & 0.304 & 0.169 & 0.268 & 0.204 & 0.305 \\ 
 & 720 & \textbf{0.193} & \textbf{0.281} & 0.199 & 0.289 & \underline{0.194} & \underline{0.288} & 0.211 & 0.302 & 0.212 & 0.304 & 0.228 & 0.320 & 0.214 & 0.307 & 0.237 & 0.314 & 0.205 & 0.304 & 0.204 & 0.301 & 0.221 & 0.319 \\ 
\addlinespace\cline{1-24} \addlinespace
\multirow[c]{4}{*}{\rotatebox{90}{ILI}} & 24 & \textbf{1.577} & \textbf{0.760} & 1.801 & 0.874 & \underline{1.783} & 0.846 & 2.086 & 0.922 & 2.182 & 1.002 & 1.804 & \underline{0.820} & 1.932 & 0.872 & 2.981 & 1.096 & 2.131 & 0.958 & 2.208 & 1.031 & 2.394 & 1.066 \\ 
 & 36 & \textbf{1.596} & \textbf{0.794} & \underline{1.743} & 0.867 & 1.746 & \underline{0.860} & 1.912 & 0.882 & 2.241 & 1.029 & 1.891 & 0.926 & 1.869 & 0.866 & 3.549 & 1.196 & 2.612 & 0.974 & 2.032 & 0.981 & 2.227 & 1.031 \\ 
 & 48 & \textbf{1.632} & \textbf{0.810} & 1.843 & 0.926 & \underline{1.716} & 0.898 & 1.985 & 0.905 & 2.272 & 1.036 & 1.752 & \underline{0.866} & 1.891 & 0.883 & 3.851 & 1.288 & 1.916 & 0.897 & 2.209 & 1.063 & 2.525 & 1.003 \\ 
 & 60 & \textbf{1.660} & \textbf{0.815} & 1.845 & 0.925 & 2.183 & 0.963 & 1.999 & 0.929 & 2.642 & 1.142 & \underline{1.831} & 0.930 & 1.914 & \underline{0.896} & 4.692 & 1.450 & 1.995 & 0.905 & 2.292 & 1.086 & 2.410 & 1.010 \\ 
\addlinespace\cline{1-24} \addlinespace
\multirow[c]{4}{*}{\rotatebox{90}{Solar}} & 96 & \textbf{0.169} & \textbf{0.195} & 0.181 & 0.247 & 0.190 & 0.244 & 0.218 & 0.235 & 0.208 & 0.255 & 0.179 & 0.232 & \underline{0.170} & 0.234 & 0.183 & \underline{0.208} & 0.198 & 0.270 & 0.199 & 0.265 & 0.381 & 0.398 \\ 
 & 192 & \textbf{0.187} & \textbf{0.207} & 0.200 & 0.259 & \underline{0.193} & 0.257 & 0.196 & \underline{0.220} & 0.229 & 0.267 & 0.201 & 0.259 & 0.204 & 0.302 & 0.208 & 0.226 & 0.206 & 0.276 & 0.220 & 0.282 & 0.395 & 0.386 \\ 
 & 336 & 0.199 & \textbf{0.213} & 0.208 & 0.269 & 0.203 & 0.266 & \underline{0.195} & \underline{0.220} & 0.241 & 0.273 & \textbf{0.190} & 0.256 & 0.212 & 0.293 & 0.212 & 0.239 & 0.208 & 0.284 & 0.234 & 0.295 & 0.410 & 0.394 \\ 
 & 720 & \textbf{0.202} & \textbf{0.216} & 0.212 & 0.275 & 0.223 & 0.281 & 0.208 & \underline{0.237} & 0.248 & 0.277 & \underline{0.203} & 0.261 & 0.215 & 0.307 & 0.215 & 0.256 & 0.232 & 0.294 & 0.243 & 0.301 & 0.377 & 0.376 \\ 
\addlinespace\cline{1-24} \addlinespace
\multirow[c]{4}{*}{\rotatebox{90}{Traffic}} & 96 & \textbf{0.360} & \textbf{0.238} & 0.368 & 0.252 & \underline{0.363} & 0.265 & 0.384 & \underline{0.250} & 0.400 & 0.280 & 0.369 & 0.257 & 0.370 & 0.262 & 0.526 & 0.288 & 0.595 & 0.312 & 0.395 & 0.275 & 0.604 & 0.330 \\ 
 & 192 & \underline{0.383} & \textbf{0.249} & \textbf{0.382} & 0.261 & 0.384 & 0.273 & 0.405 & \underline{0.257} & 0.412 & 0.288 & 0.400 & 0.272 & 0.386 & 0.269 & 0.503 & 0.263 & 0.613 & 0.322 & 0.407 & 0.280 & 0.610 & 0.338 \\ 
 & 336 & \underline{0.395} & \textbf{0.259} & \textbf{0.393} & 0.268 & 0.396 & 0.277 & 0.424 & \underline{0.265} & 0.426 & 0.301 & 0.407 & 0.272 & 0.396 & 0.275 & 0.505 & 0.276 & 0.626 & 0.332 & 0.417 & 0.286 & 0.626 & 0.341 \\ 
 & 720 & \textbf{0.435} & \textbf{0.278} & 0.438 & 0.297 & 0.445 & 0.308 & 0.452 & \underline{0.283} & 0.478 & 0.339 & 0.461 & 0.316 & \underline{0.435} & 0.295 & 0.552 & 0.301 & 0.635 & 0.340 & 0.454 & 0.308 & 0.643 & 0.347 \\ 
\addlinespace\cline{1-24} \addlinespace
\multicolumn{2}{c|}{$1^{st}$ Count}& \textbf{30} & \textbf{38} & \underline{3} & 0 & 0 & 0 & 0 & \underline{1} & 2 & \underline{1} & 1 & 0 & 1 & 0 & 2 & 0 & 0 & 0 & 1 & 0 & 0 & 0 \\ 
\addlinespace

\bottomrule
\end{tabular}
}
\end{table*}

\begin{table}[t!]
\caption{The comparsion among different distance metrics. Results are averaged from all forecasting horizons. }
\resizebox{1\columnwidth}{!}{
\footnotesize
\begin{tabular}{@{}cc|cc|cc|cc|cc|cc}
\toprule
  \multicolumn{2}{c|}{\textbf{Models}} & \multicolumn{2}{c}{\textbf{DUET}} & \multicolumn{2}{c}{\textbf{Euclid}} & \multicolumn{2}{c}{\textbf{Cosine}} & \multicolumn{2}{c}{\textbf{DTW}} & \multicolumn{2}{c}{\textbf{Random Mask}}  \\
  \addlinespace\cline{1-12} \addlinespace
\multicolumn{2}{c|}{Metrics}  & mse & mae & mse & mae & mse & mae & mse & mae & mse & mae   \\
\midrule
\multicolumn{2}{c|}{ETTh2} &  \textbf{0.334}&  	\textbf{0.383}&  	0.343&  	0.390&  	0.343&  	0.391&  	0.344&  	0.391&  	0.341&  	0.389  \\
 \addlinespace\cline{1-12} \addlinespace
\multicolumn{2}{c|}{ETTm2} & \textbf{0.247}& 	\textbf{0.307}& 	0.255& 	0.313& 	0.256& 	0.312& 	0.256& 	0.314& 	0.257& 	0.315  \\
 \addlinespace\cline{1-12} \addlinespace
\multicolumn{2}{c|}{Weather} & \textbf{0.218}& 	\textbf{0.252}& 	0.224 &	0.257 &	0.225 &	0.257 &	0.228 &	0.262 &	0.227 &	0.259  \\
 \addlinespace\cline{1-12} \addlinespace
\multicolumn{2}{c|}{Traffic} & \textbf{0.393}& 	\textbf{0.256}& 	0.397& 	0.260	& 	0.398& 0.262& 	0.396 &	0.259& 	0.397 &	0.261
 \\
\addlinespace
\bottomrule
\end{tabular}
}

\label{The comparsion between different distance metric avg.}
\end{table}

\subsection{Main Results}
Comprehensive forecasting results are listed in Table~\ref{Common Multivariate forecasting results avg} with the best in bold and the second underlined. We have the following observations: \par
1) Compared with forecasters of different structures, DUET achieves an excellent predictive performance. From the perspective of absolute performance, DUET demonstrates a significant improvement against the second-best baseline PDF, with an impressive 7.1\% reduction in MSE and a 6.5\% reduction in MAE. \par
2) Considering different channel strategies, DUET also shows the advantage of CSC strategy. Compared with CI models such as FITS and PatchTST, DUET outperforms them comprehensively, especially on large datasets with abundant channel correlations such as Solar and Traffic. Besides, as the previous state-of-the-art, iTransformer and Crossformer with the CD strategy fail in many cases of ETT~(4 subsets). Due to the weak correlations among channels of ETT, they may be affected by noise channels~(uncorrelated channels), leading to a decrease in performance. In contrast, DUET with CSC strategy implements a soft clustering among channels (where each channel only focuses on the channels related to it), can better address this situation. 
\par
3) DUET also demonstrates strong performance in tackling temporal heterogeneity caused by Temporal Distribution Shift. Compared to the state-of-the-art Non-stationary Transformer model for non-stationary time series modeling. DUET achieves a significant reduction of 32.4\% in MSE and 21.7\% in MAE. This result indicates that the proposed Temporal Clustering Module is more effective in modeling heterogeneous temporal patterns.

\subsection{Model Analyses}
\subsubsection{Ablation Studies}
\label{sec: abl}
To ascertain the impact of different modules within DUET, we perform ablation studies focusing on the following components. 

\noindent
(1) \textit{w/o TCM:} Remove the Temporal Cluster Module.

\noindent
(2) \textit{w/o CCM:} Remove the Channel Cluster Module.

\noindent
(3) \textit{Full Attention:} Replace the Mask Attention with Full Attention.

\noindent
(4) \textit{Tem-Info:} Change the channel distance calculation from the frequency domain to the temporal domain.

Table~\ref{tab: Ablation study average results} illustrates the unique impact of each module. We have the following observations: 1) When the temporal clustering module is removed, the performance decreases relatively slightly for datasets with fewer distribution changes such as Traffic. While evaluating datasets with substantial temporal distribution changes, such as ETTh2, the performance drops more significantly. This indicates the effectiveness of the temporal clustering module. 2) When the channel clustering module is removed and our model works in CI strategy, the performance on Traffic, which has strong inter-channel correlations, drops significantly. This highlights the necessity of soft clustering among channels. 3) When a full attention mechanism~(where each variable calculates attention scores with all other variables) is used among channels instead of a masked matrix, our model works in CD strategy and the performance on the ETTh2 dataset, which has less noticeable inter-channel correlations, also drops significantly. This demonstrates the effectiveness of our proposed CSC strategy with the masked matrix applying to channels. 4) When the distances among channels are calculated in the temporal domain instead of the frequency domain, the algorithm's performance also decreases. This suggests that clustering channels in the frequency domain introduces more semantic information and helps learn a robust distance metric to evaluate the channel relations.

\subsubsection{Comparsion Among Different Distance Metrics}
To obtain the mask matrix in the channel view, we adopt a learnable Mahalanobis distance metric to adaptively evaluate the correlations among channels in the frequency domain. We have demonstrated in \ref{sec: abl} that frequency domain actually introduces more useful information for channel clustering. Then, we demonstrate that the learnable Mahalanobis distance metric is better than other similarity measurement methods. We choose to replace the learnable metric with non-learnable similarity measurement methods: Euclidean distance, Cosine similarity, and DTW~\cite{muller2007dynamic}. As shown in Table~\ref{The comparsion between different distance metric avg.}, using these methods results in a decrease in model performance, which fully demonstrates the effectiveness of the learnable metric. Furthermore, we also replace the learned mask matrix with a random mask matrix and prove it performs far worse than the learned one.

\subsubsection{Parameter Sensitivity: Varying the Number of Extractors}
DUET clusters time series into $M$ classes based on their temporal distributions, where $M$ is the size of Linear-based Pattern Extractor Cluster. We analyze the influence of different $M$ values on prediction accuracy in Table~\ref{Parameter sensitivity study}. We have the following observations: 1) The performance of M=1 is inferior to that of M not equal to 1. 2) Datasets from the same domain, such as the electricity domain datasets ETTh1 and ETTh2, their best values of $M$ are the same, which are 4. 3) While for datasets from different domains, such as ILI (health) and Exchange (economic), their best values of $M$ are 2 and 5. 4) With the most proper $M$, the performance leads other variants  remarkablely in most cases. This highlights the effectiveness of clustering in the temporal view, and implies datasets from the same domain often have similar temporal distributions, vice versa.

\renewcommand{\arraystretch}{1}
\begin{table}[t]
\caption{Parameter sensitivity study. The prediction accuracy varies with M which is the number of extractors. Results are averaged from all forecasting horizons. }
\resizebox{1\columnwidth}{!}{
\footnotesize
\begin{tabular}{cc|cccccccccc}
\toprule
  \multicolumn{2}{c|}{\textbf{
Extractors}} & \multicolumn{2}{c}{\textbf{M=1}} & \multicolumn{2}{c}{\textbf{M=2}} & \multicolumn{2}{c}{\textbf{M=3}} & \multicolumn{2}{c}{\textbf{M=4}} & \multicolumn{2}{c}{\textbf{M=5}}  \\
\multicolumn{2}{c|}{Metrics}  & mse & mae & mse & mae & mse & mae & mse & mae & mse & mae   \\

\midrule
\multicolumn{2}{c|}{ETTh1}& 0.402	&0.421	&0.408&	0.426&	0.403&	0.424&	\textbf{0.398}	&\textbf{0.418}&	0.411&	0.430\\
 \addlinespace\cline{1-12} \addlinespace
\multicolumn{2}{c|}{ETTh2}&  0.346	&0.389	&0.343&	0.388&	0.342	&0.387	&\textbf{0.334}	&\textbf{0.383}&	0.341&	0.387  \\
 \addlinespace\cline{1-12} \addlinespace
\multicolumn{2}{c|}{ILI}&1.666&	0.798	&\textbf{1.616}	&\textbf{0.795}	&1.701	&0.816&	1.708	&0.818	&1.679	&0.810\\
 \addlinespace\cline{1-12} \addlinespace 
\multicolumn{2}{c|}{Exchange}& 0.365&	0.404&	0.348&	0.397	&0.366&	0.405	&0.322	&0.383&	\textbf{0.280}	&\textbf{0.364}
 \\
\addlinespace
\bottomrule
\end{tabular}
}
\label{Parameter sensitivity study}
\end{table}

\begin{figure}[t!]
  \centering
  \includegraphics[width=1\linewidth]{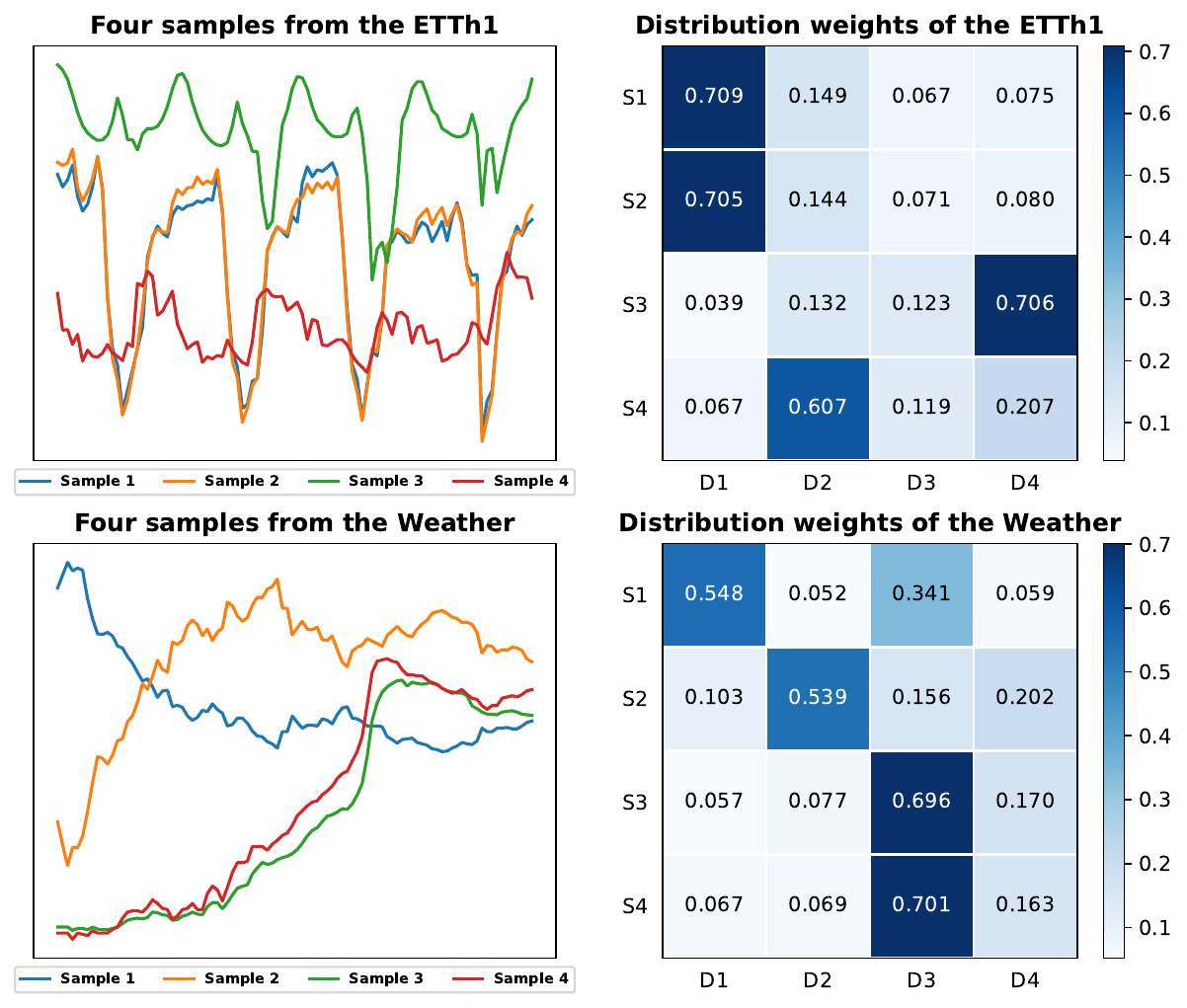}
  \caption{The distribution weights of different time series samples from the ETTh1 and Weather. S1-S4 denote distinct samples, while D1-D4 represent different distributions.}
    \label{fig:pick temporal}
\end{figure}

\begin{figure}[t]
  \centering
  \includegraphics[width=\linewidth]{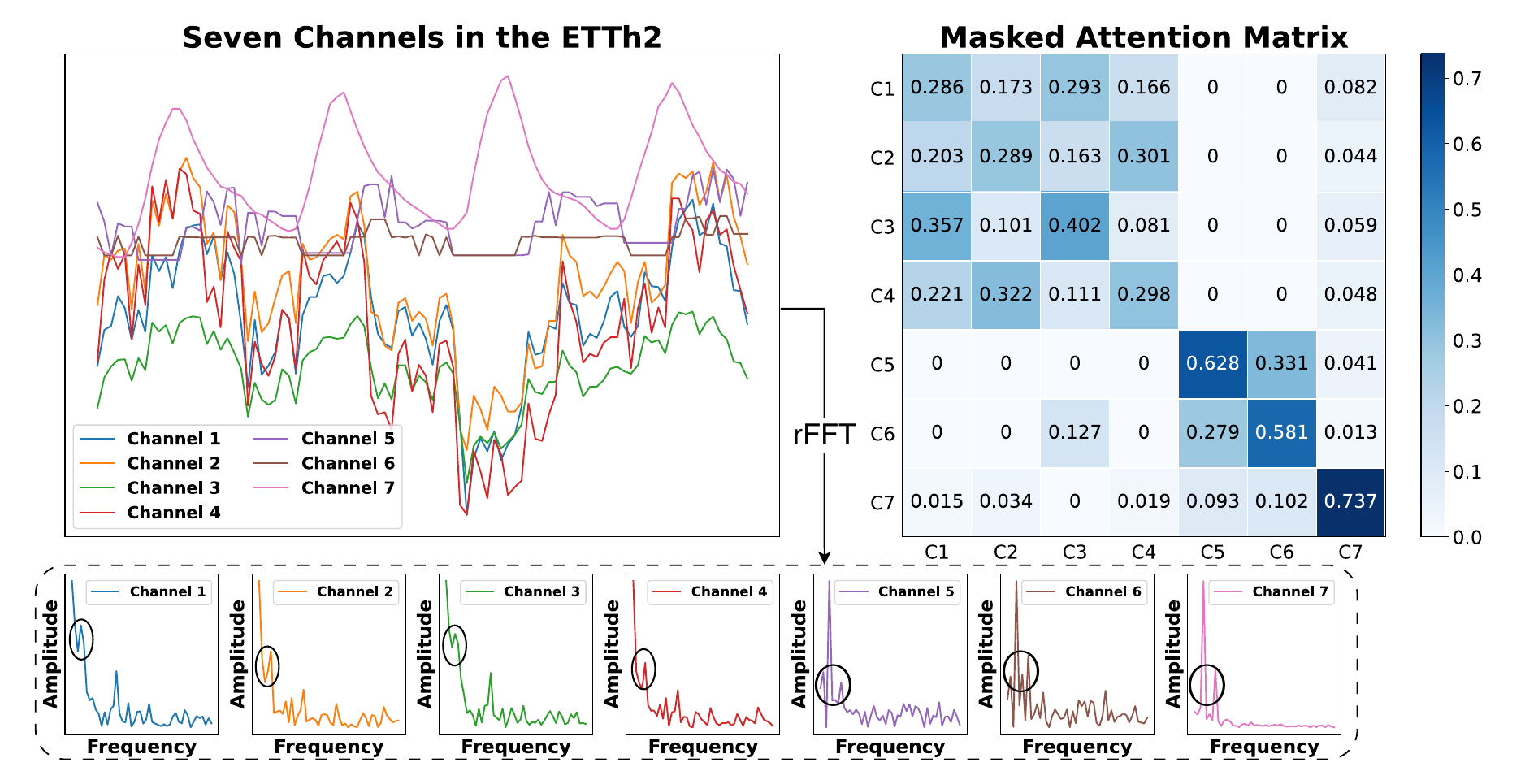}
  \caption{The attention scores of different channels from the ETTh2. C1--C7 denote distinct channels.}
    \label{fig:pick channel}
\end{figure}

\subsubsection{Visualization of Distributions Weights:}
The distribution router in TCM is designed to cluster samples by extracting the mean and variance of the implict distributions. During clustering, we do not consider the temporal alignment of the samples; instead, we focus on the distributional characteristics of each sample to determine its distribution category. We show four time series samples with the same length of 96 from the ETTh1 and Weather datasets respectively, and described their unique distribution weights in Figure~\ref{fig:pick temporal}. We have the following observations: 1) Samples 1 and 2 in ETTh1 exhibit similar seasonal patterns, show similar distribution weights. However, for samples 1, 3, and 4, their distributions are vastly different. 2) Sample 3 and Sample 4 in Weather have similar trends, show similar distribution weights. However, for samples 1, 2, and 3, their distributions are vastly different. These observations emphasize the adaptability of DUET and its ability to distinguish and cluster samples with different distributions.

\subsubsection{Visualization of Channel Weights:}
We show the mutual attention weights of the seven channels in ETTh2 to exhibit the effectiveness of our Channel Clustering Module (CCM). For channels in the same time interval, the CCM measures their correlations for the downstream prediction task in the frequency space and generates corresponding mask matrices to obtain a clustering effect. Figure~\ref{fig:pick channel} shows an example of this flexible clustering paradigm by visualizing the masked attention weights in the Fusion Module. We can observe that channels with similar frequency component combinations may be clustered into a soft group while the cross-group correlations are also partially kept---(see 0.127 between C6 and C3) to maximize the neighbor information gain for the prediction task.

\section{Conclusions}
In this paper, we propose a general framework, DUET, which introduces a dual clustering on the temporal and channel dimensions to enhance multivariate time series forecasting. It integrates a Temporal Clustering Module (TCM) which clusters time series into fine-grained distributions. Various pattern extractors are then designed for different distribution clusters to capture their unique temporal patterns, modeling the heterogeneity of temporal patterns. Furthermore, we intorduce the Channel Clustering Module~(CCM) using a channel-soft-clustering strategy. This captures the relationships among channels in the frequency domain through metric learning and applies sparsification. Finally, the Fusion Module (FM), based on a masked attention mechanism, efficiently combines the temporal features extracted by the TCM with the channel mask matrix generated by the CCM. These innovative mechanisms collectively empower DUET to achieve outstanding prediction performance.

\begin{acks}
This work was partially supported by National Natural Science Foundation of China (62472174, 62372179). Jilin Hu is the corresponding author of the work.
\end{acks}

\clearpage
\bibliographystyle{ACM-Reference-Format}
\bibliography{sample-base}

\clearpage
\appendix
\section{THEORETICAL ANALYSIS}
\label{app}
\subsection{Computational Complexity Analysis}
We compare the theoretical complexity across different Transformer-based models in Table~\ref{tab:computational complexity} because the main Computational complexity of DUET comes from the Transformer-based Fusion Module. Subsequent works mainly utilize attention mechanism to capture temporal dependencies and channel correlations. Benefiting from our dual clustering design, DUET efficiently extracts temporal patterns so that we only need to capture the correlations among channels with attention mechanism, which can be paralleled on the look-back window size. This ensures low computational costs even when the look-back window size is extremely large.

\subsection{Noisy Gating}
\label{app:noisygating}
We combine the reparameterization trick and original Noisy Gating technique in following formulas of our paper:
\begin{align*}    
Z_n &=\textit{Encoder}_{\mu}(X_{n,:}) + \epsilon \odot \textit{Softplus}(\textit{Encoder}_{\sigma}(X_{n,:}))\\
H(X_{n,:}) &= W^H \cdot Z_n
\end{align*}

We provide evidence that the combination is computational equivalent to the original Noisy Gating. First, we simplify the symbolic representation for legibility. For our formulas, we simplify them as follows:
\begin{align*}    
H(X_{n,:}) &= W^H \cdot (\mu + \epsilon \odot \sigma),
\end{align*}
where $\mu,\sigma,\epsilon \in \mathbb{R}^{M}$ and $ \epsilon_i \sim \mathcal{N}(0,1)$. Then we formualize the original Noisy Gating technique in our context:
\begin{align*}    
H^\prime(X_{n,:}) &= W^{H_1}\cdot (\mu + \epsilon \odot \sigma) + \epsilon^\prime \odot \textit{Softplus}(W^{H_2}\cdot (\mu + \epsilon \odot \sigma)),
\end{align*}
where $\epsilon_i, \epsilon^\prime_i \sim \mathcal{N}(0,1)$ are all \textbf{i.i.d} and independent of each other. Next, we show the equivalency between the two formulas. We first simplify the part $\epsilon^\prime \odot\textit{Softplus}(W^{H_2}\cdot (\mu + \epsilon \odot \sigma))$ as $\epsilon^\prime \odot V^\prime$, where $V^\prime \in \mathbb{R}^M$ is a learnable vector. Then we consider the every vector component in this formula:
\begin{align*}
    H^\prime(X_{n,:}) &= W^{H_1}\cdot (\mu + \epsilon \odot \sigma) + \epsilon^\prime \odot V^\prime\\
    H^\prime(X_{n,:})_i &= (W^{H_1} \cdot \mu)_i + (W^{H_1} \cdot (\epsilon \odot \sigma))_i + V^\prime_i \cdot \epsilon^\prime_i\\
    H^\prime(X_{n,:})_i &= (W^{H_1} \cdot \mu)_i + \sum_{j=1}^M W^{H_1}_{ij}\cdot \sigma_j \cdot \epsilon_j + \epsilon^\prime_i \cdot V^\prime_i
\end{align*}
the $H^\prime(X_{n,:})_i$ obeys normal distribution:
\begin{align*}
H^\prime(X_{n,:})_i &\sim \mathcal{N}((W^{H_1} \cdot \mu)_i,\sum_{j=1}^M (W^{H_1}_{ij}\cdot \sigma_j)^2 + (V^\prime_i)^2)
\end{align*}
for learnable components $V^\prime$ and $W^{H_1}$, they can be adaptively tuned and approximately equivalent to:
\begin{align*}
    H^\prime(X_{n,:})_i &\sim \mathcal{N}((\hat{W}^{H_1} \cdot \mu)_i,\sum_{j=1}^M (\hat{W}^{H_1}_{ij}\cdot \sigma_j)^2)\\
    e.g. \ (\hat{W}^{H_1}_{ij})^2 &= (W^{H_1}_{ij})^2 + \frac{(V^\prime_i)^2}{(M \cdot \sigma_j)^2}
\end{align*}
while $H(X_{n,:})_i$ also obeys the same form of distribution:
\begin{align*}
        H(X_{n,:})_i &\sim \mathcal{N}((W^H \cdot \mu)_i,\sum_{j=1}^M (W^H_{ij}\cdot \sigma_j)^2)
\end{align*}
Q.E.D.

\begin{figure*}[t!]
  \centering
  \includegraphics[width=1\linewidth]{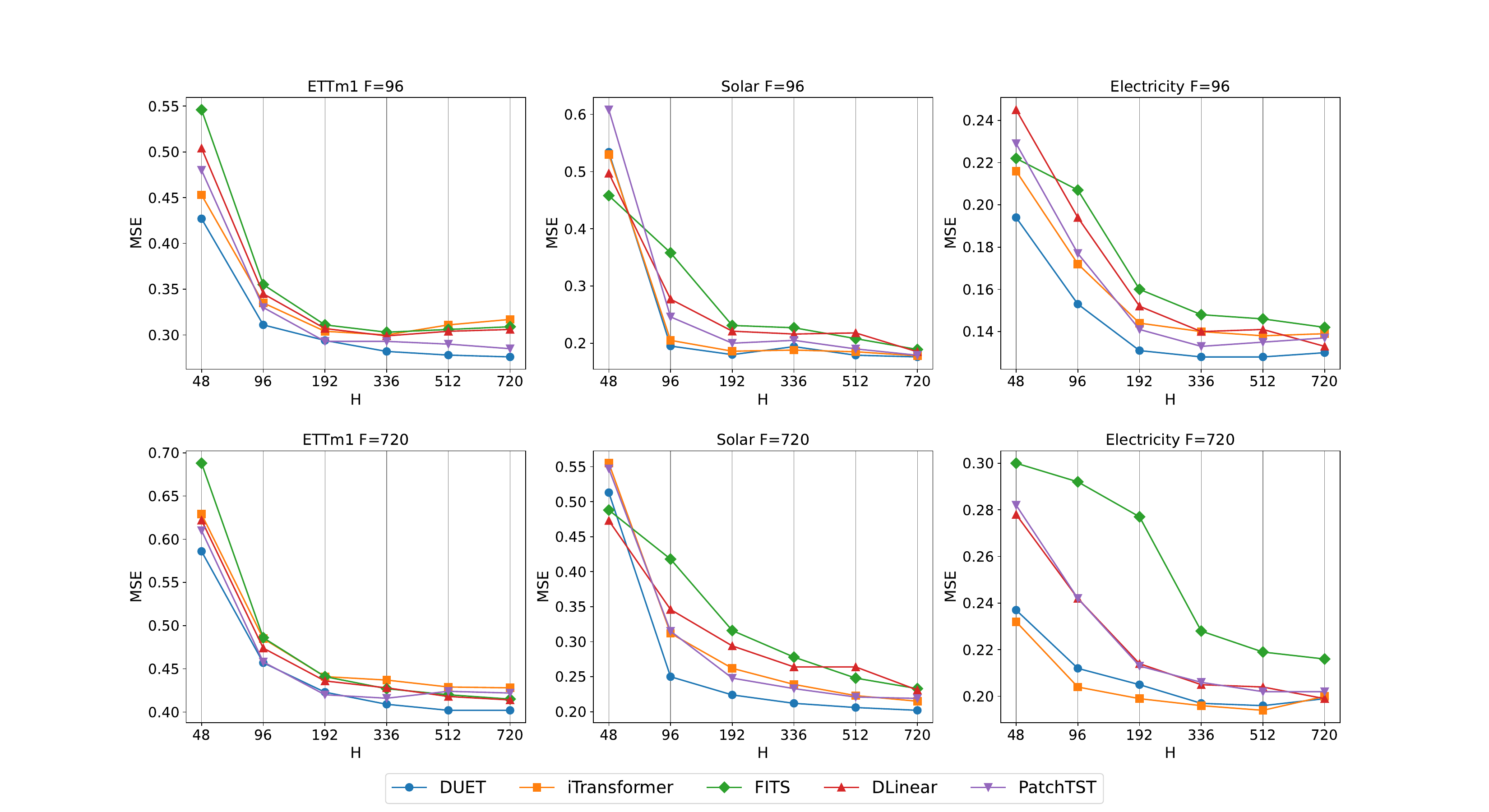}
  \caption{Forecasting performance (MSE) with varying look-back windows on 3 datasets: ETTm1, Solar, and Electricity. The look-back windows are selected to be H = 48, 96, 192, 336, 512, and 720, and the forecasting horizons are F = 96, and 720.}
    \label{fig:look_back_line_chart}
\end{figure*}

\subsection{Fourier Transform in $L^2$ Space}
\label{app:fourier}
We introduce the Fourier transform from the perspective of functional analysis: 
Taking \( L^2(-\pi,\pi) \) as an example, where the inner product is defined as \( (f,g) = \int_{E} fg \, d\mu \) for \( f, g \in L^2(-\pi,\pi) \), induced \( L^2 \)-norm \( (\int_{E} |f|^2 \, d\mu)^{\frac{1}{2}} \) (conforming to the inner product definition norm), this space constitutes a Hilbert space. Since functions eligible for Fourier expansion must satisfy conditions of Riemann integrability, considering only such functions reduces the inner product to the Riemann integral \( (f,g) = \int_{-\pi}^{\pi} f(x)g(x) \, dx \). Direct computation confirms that the trigonometric function series:
\begin{align*}
     [e_n] = \left[\frac{1}{\sqrt{2\pi}}, \frac{1}{\sqrt{\pi}} \cos(nx), \frac{1}{\sqrt{\pi}} \sin(nx), \ldots, \frac{1}{\sqrt{\pi}} \cos(nx), \ldots \right] 
\end{align*}
forms a standard orthonormal system, and \( \text{span}[e_n] \) is dense in \( L^2(-\pi,\pi) \), hence serving as a standard orthonormal basis. It is evident that Fourier coefficients can be expressed as:
\begin{align*}
&\begin{cases}
\frac{a_0}{2} = \frac{1}{2\pi} \int_{-\pi}^{\pi} f(x) \, dx = \frac{1}{\sqrt{2\pi}} (f, \frac{1}{\sqrt{2\pi}}) \\
a_n = \frac{1}{\pi} \int_{-\pi}^{\pi} \cos(nx) f(x) \, dx = \frac{1}{\sqrt{\pi}} (f, \frac{1}{\sqrt{\pi}} \cos(nx)) \quad \\
b_n = \frac{1}{\pi} \int_{-\pi}^{\pi} \sin(nx) f(x) \, dx = \frac{1}{\sqrt{\pi}} (f, \frac{1}{\sqrt{\pi}} \sin(nx)) \quad
\end{cases}
\\
&(n=1,2,3,\ldots)
\end{align*}
thus yielding:
\begin{align*}
    &f(x) = (f, \frac{1}{\sqrt{2\pi}}) \frac{1}{\sqrt{2\pi}} + \\ &\sum_{n=1}^{\infty} \left[ (f, \frac{1}{\sqrt{\pi}} \cos(nx)) \frac{1}{\sqrt{\pi}} \cos(nx) + (f, \frac{1}{\sqrt{\pi}} \sin(nx)) \frac{1}{\sqrt{\pi}} \sin(nx) \right]
\end{align*}
From this point, Fourier Transform actually plays as an orthogonal transformation in $L^2$ space, thus $a_n$ and $b_n$ are coordinates under the trigonometric function basis and can be seen as a point in the space.

\begin{table}[t!]
    \caption{Theoretical computational complexity per layer in Transformer-based models. $H$ denotes the look-back window size, $N$ denotes the number of channels, $p$ denotes the patch size in patch-based methods, and $max(p_i)$ denotes the maximum decoupled periodic length of PDF.}
    \centering
    \footnotesize
    \begin{tabular}{@{}c|c}
    \toprule
         Method& Time complexity  \\
    \midrule
        Autoformer & $O(H \log H)$\\
    \midrule
        PatchTST & $O((\frac{H}{p})^2)$\\
    \midrule
        Crossformer & $O((\frac{H}{p})^2 + N^2)$\\
    \midrule
        Pathformer & $O(H \log H)$\\
    \midrule
        iTransformer & $O(N^2)$\\
    \midrule
        PDF & $O((\frac{max(p_i)}{p})^2)$\\
    \midrule
        DUET (Ours) & $O(N^2)$\\
    \bottomrule
    \end{tabular}
    \label{tab:computational complexity}
\end{table}

\section{HYPERPARAMETER SENSITIVITY}
\subsection{Varying Look-back Window}
In time series forecasting tasks, the size of the look-back window determines how much historical information the model receives. We select models with better predictive performance from the main experiments as baselines. We configure different look-back window to evaluate the effectiveness of DUET and visualize the prediction results for look-back window H of 48, 96, 192, 336, 512, 720, and the forecasting horizons are F = 96, 720. From Figure ~\ref{fig:look_back_line_chart}, DUET consistently outperforms the baselines on the ETTm1, Solar, and Electricity. As the look-back window increases, the prediction metrics of DUET continue to decrease, indicating that it is capable of modeling longer sequences.

\subsection{The Advantages of Dual Clustering}

\begin{figure*}[!htbp]
  \centering
  \includegraphics[width=0.65\linewidth]{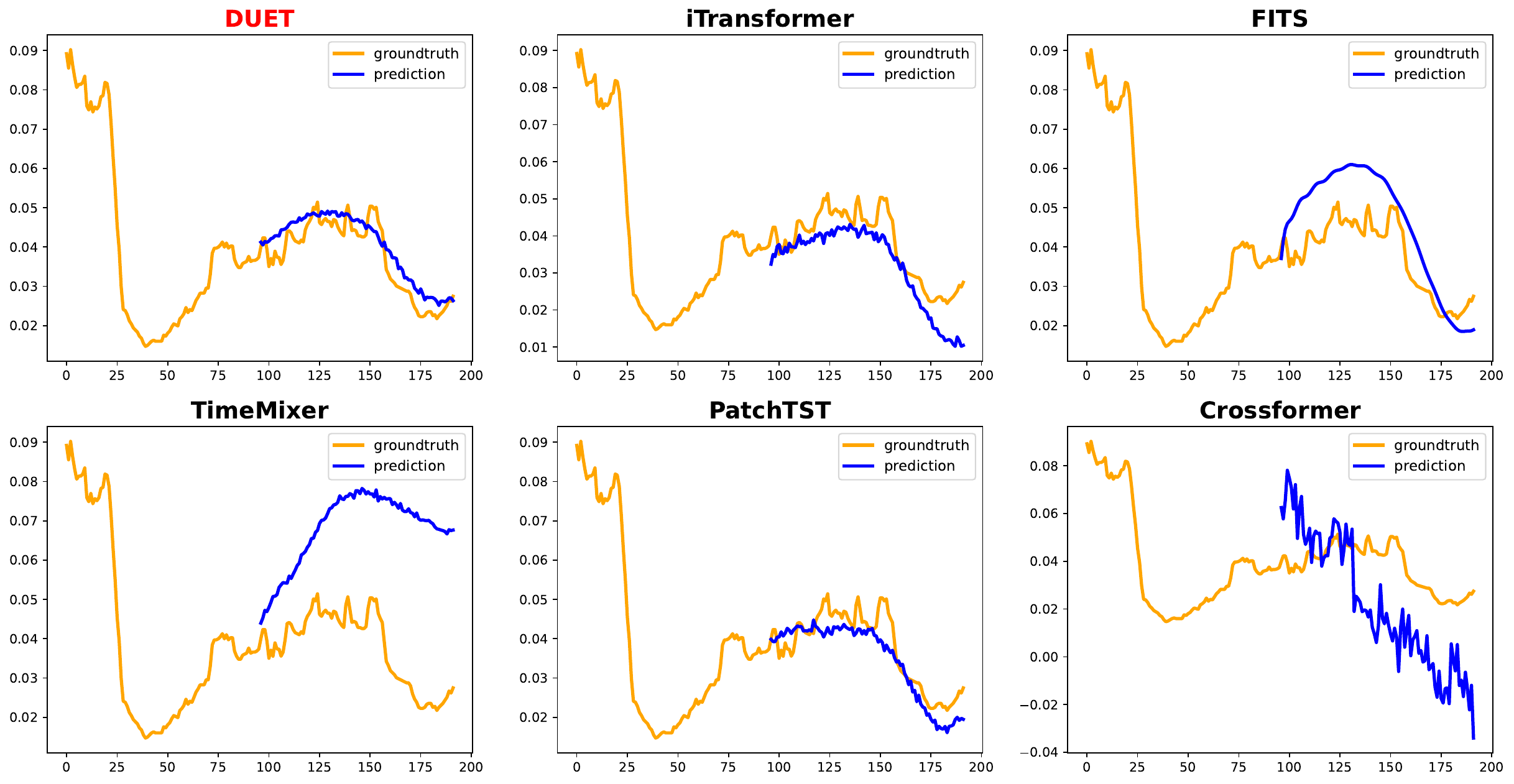}
  \caption{Visualization of input-96-predict-96 results on the Weather dataset.}
  \label{fig: weather}
\end{figure*}

\begin{figure*}[!htbp]
  \centering
  \includegraphics[width=0.65\linewidth]{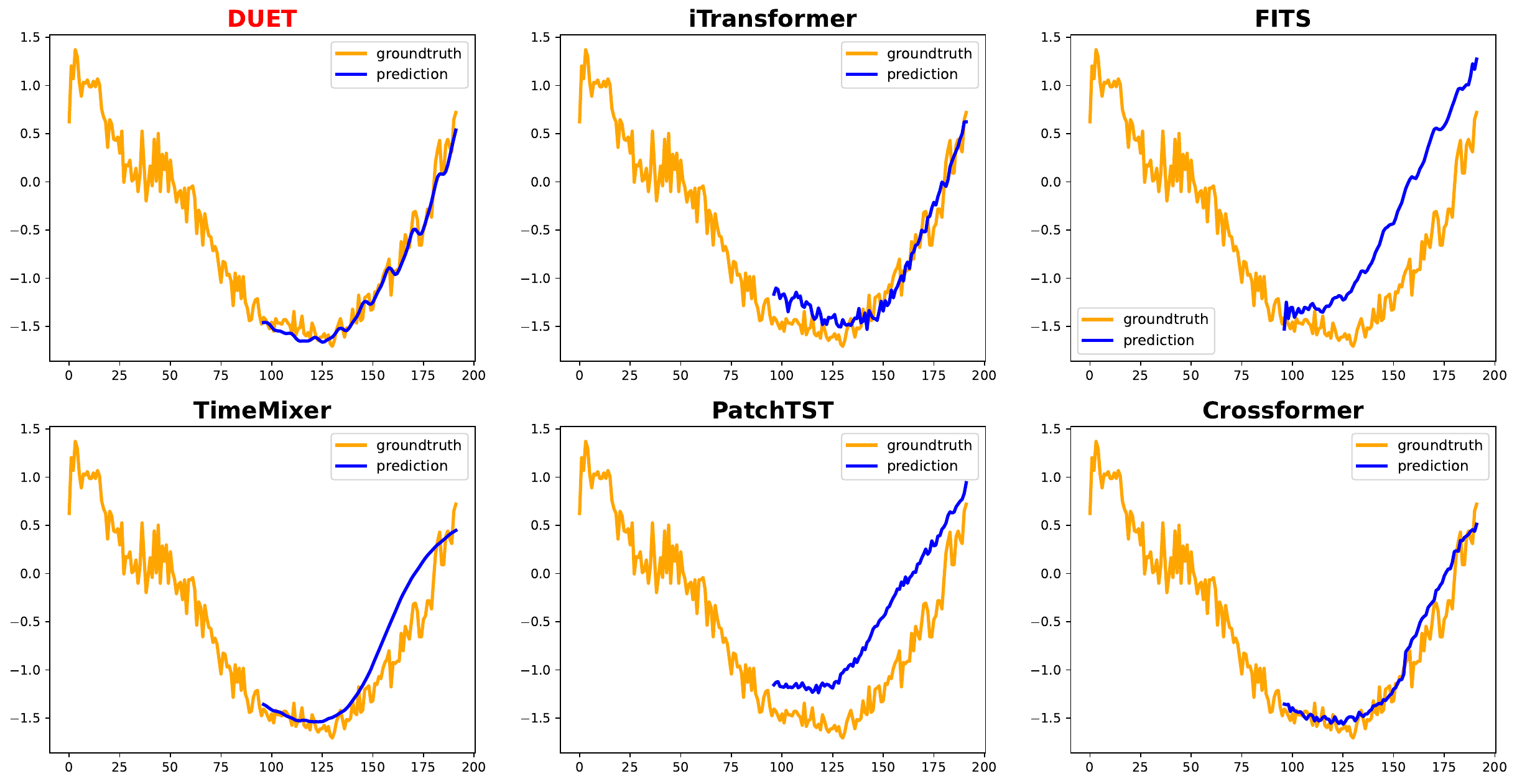}
  \caption{Visualization of input-96-predict-96 results on the PEMS04 dataset.}
  \label{fig: pems04}
\end{figure*}

As aforementioned, recent approaches assume the time series is sampled from a single distribution and use a single backbone to model the heterogeneous temporal patterns. While DUET clusters time series into different distributional categories and use parameter independent backbones to model heterogeneous temporal patterns. Figure~\ref{fig: weather} and Figure~\ref{fig: pems04} show that other baselines perform insufficient sensibility to distributional changing while DUET fits well to such cases. For datasets like Electricity with strong seasonality, CI strategy fails to take advantage of correlations among channels while CD strategy suffers from the adverse effects of delays among channels. Both of them capture a fuzzy seasonality while DUET with the CSC strategy precisely leverages pros and cons from other channels and shows a more accurate performance.

\begin{figure*}[!htbp]
  \centering
  \includegraphics[width=0.65\linewidth]{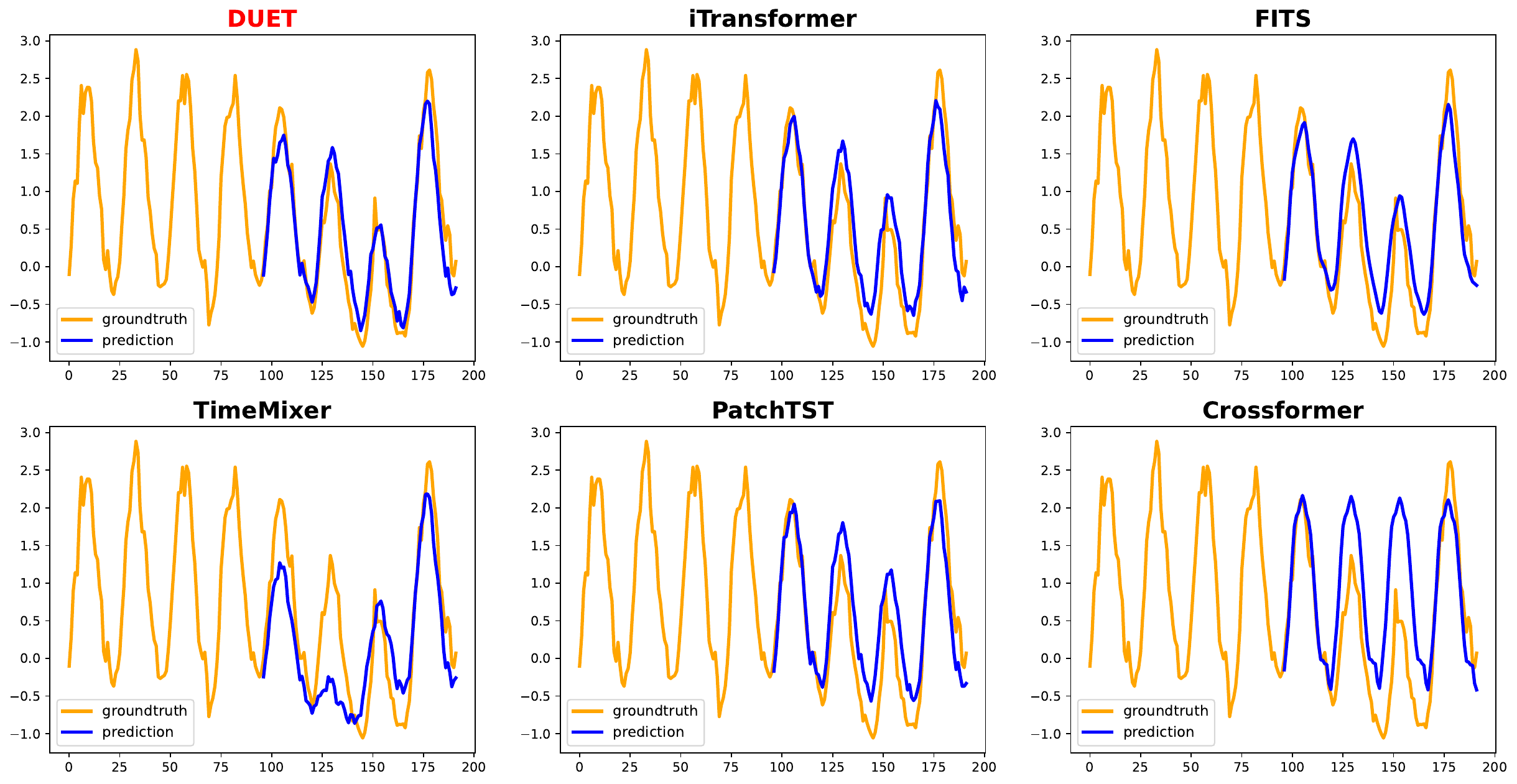}
  \caption{Visualization of input-96-predict-96 results on the Electricity dataset.}
  \label{fig: ecl}
\end{figure*}

\end{document}